\newcommand{\dataset}{\mathcal{D}}
\newcommand{\footage}{{v}}
\newcommand{\edit}{{x}}
\newcommand{\score}{s}
\newcommand{\scoresum}{S}
\newcommand{\feedback}{f}

\newcommand{\reward}{r}
\newcommand{\model}{\mathcal{M}}
\newcommand{\tabtitlecolor}{\rowcolor{blue!6}}

\documentclass{article}

\usepackage{microtype}
\usepackage{graphicx}
\usepackage{subcaption}
\usepackage{booktabs} 
\usepackage{colortbl}
\usepackage{graphicx}
\usepackage{booktabs}
\usepackage{amsmath} 
\usepackage{amssymb}
\usepackage{mathtools}
\usepackage{amsthm}
\usepackage{multirow}
\usepackage{multicol}
\usepackage{mathrsfs}
\usepackage{diagbox}
\usepackage{dsfont}
\usepackage{hyperref}
\usepackage{xurl} 
\usepackage{enumitem}

\usepackage{hyperref}

\usepackage[accepted]{icml2026}

\usepackage{amsmath}
\usepackage{amssymb}
\usepackage{mathtools}
\usepackage{amsthm}

\usepackage{xcolor} 
\usepackage{soul}   
\usepackage{rotating}   
\usepackage{makecell} 
\usepackage{tcolorbox}
\tcbuselibrary{breakable, skins} 
\usepackage{xcolor}

\usepackage{enumitem}
\usepackage{amssymb}
\usepackage{framed}
\usepackage{xcolor}
\usepackage{hyperref}
\hypersetup{
    colorlinks=true,
    linkcolor=blue,
    filecolor=magenta,      
    urlcolor=cyan,
}

\definecolor{dimcolor}{RGB}{50, 100, 150}

\newenvironment{scorebox}[1]{%
    
\begin{framed}
    \noindent\textbf{#1}
    \vspace{5pt}
    \hrule
    \vspace{5pt}
}{%
    \end{framed}
}

\usepackage[capitalize,noabbrev]{cleveref}

\theoremstyle{plain}

\theoremstyle{definition}

\theoremstyle{remark}

\usepackage[textsize=tiny]{todonotes}

\usepackage{xcolor}
\usepackage{xspace}
\newcommand{\ie}{\textit{i.e.},\xspace}

\definecolor{MyDarkRed}{rgb}{0.8,0.02,0.02}
\definecolor{MyDarkBlue}{rgb}{0.02,0.02,0.8}
\definecolor{MyDarkGreen}{rgb}{0.02,0.5,0.02}
\definecolor{darkgreen}{rgb}{0.0, 0.5, 0.0}

\icmltitlerunning{VlogReward: Learning Multi-Dimensional Evaluation for Vlog Editing}

\begin{document}

\twocolumn[
  \icmltitle{VlogReward: Learning Multi-Dimensional Evaluation for Vlog Editing}

  \icmlsetsymbol{intern}{‡}
  \icmlsetsymbol{corresponding}{*}

  \begin{icmlauthorlist}
    \icmlauthor{Yexiang Liu}{intern,CASIA,UCAS,bytedance}
    \icmlauthor{Wen Zhong}{bytedance}
    \icmlauthor{Sijie Zhu}{bytedance}
    \icmlauthor{Xin Gu}{bytedance}
    \icmlauthor{Fan Chen}{bytedance}
    \icmlauthor{Junxian Duan}{CASIA,UCAS}
    \icmlauthor{Jie Cao}{CASIA,UCAS}
    
    \icmlauthor{Longyin Wen}{bytedance}
    \icmlauthor{Zhenfang Chen}{corresponding,bytedance}
  \end{icmlauthorlist}

  \vspace{5pt}
  \centering
  Project Page: \url{https://VlogReward.github.io}

  \icmlaffiliation{CASIA}{MAIS \& NLPR, Institute of Automation, Chinese Academy of Sciences, Beijing, China}
  \icmlaffiliation{UCAS}{University of Chinese Academy of Sciences, Beijing, China}
  \icmlaffiliation{bytedance}{ByteDance Intelligent Creation, San Jose, USA}

  \icmlcorrespondingauthor{Yexiang Liu}{liuyexiang2023@ia.ac.cn} 
  \icmlcorrespondingauthor{Zhenfang Chen}{zhenfang.chen@bytedance.com}
    
  \icmlkeywords{Machine Learning, ICML}
  \vskip 0.3in
]



\printAffiliationsAndNotice{‡\,This work was done during the internship at ByteDance. *\,Corresponding author.}  

\begin{abstract}
  The rapid rise of vlogs as a personalized storytelling medium has created a demand for automated systems to evaluate and refine vlog editing plans. However, vlog assessment is highly subjective and remains challenging due to a lack of standardized criteria, dataset and benchmark, and effective reward models. To address these challenges, we define a comprehensive vlog evaluation framework guided by professional vlog creators and product managers, establishing a taxonomy of six key dimensions, \textit{i.e.}, \textit{Creativity}, \textit{Consistency}, \textit{Concept Design}, \textit{Cinematography}, \textit{Narration}, and \textit{Pacing}. Subsequently, we curate a large-scale dataset of 100k vlog edits and a dedicated benchmark, \textbf{VRMBench}, to evaluate the vlog rewarding capabilities of Multimodal Large Language Models (MLLMs). Finally, we present \textbf{VlogReward}, a robust vlog reward model that can provide both fine-grained multi-dimensional scores and actionable feedback for iterative refinement.
  Technically, we enhance the Group Relative Policy Optimization (GRPO) framework by introducing an adjustable inter-group comparison reward, which mitigates the ``direction blindness'' issue of standard GRPO and enables the model to better distinguish varied-quality edits.
  VlogReward achieves state-of-the-art results that significantly outperform existing MLLMs, including GPT-5 and Gemini-3-Pro. 
  We hope that our study can help vlog creators and foster automated vlog evaluation and refinement systems.

\end{abstract}

\begin{figure}
    \centering
    \includegraphics[width=1\linewidth]{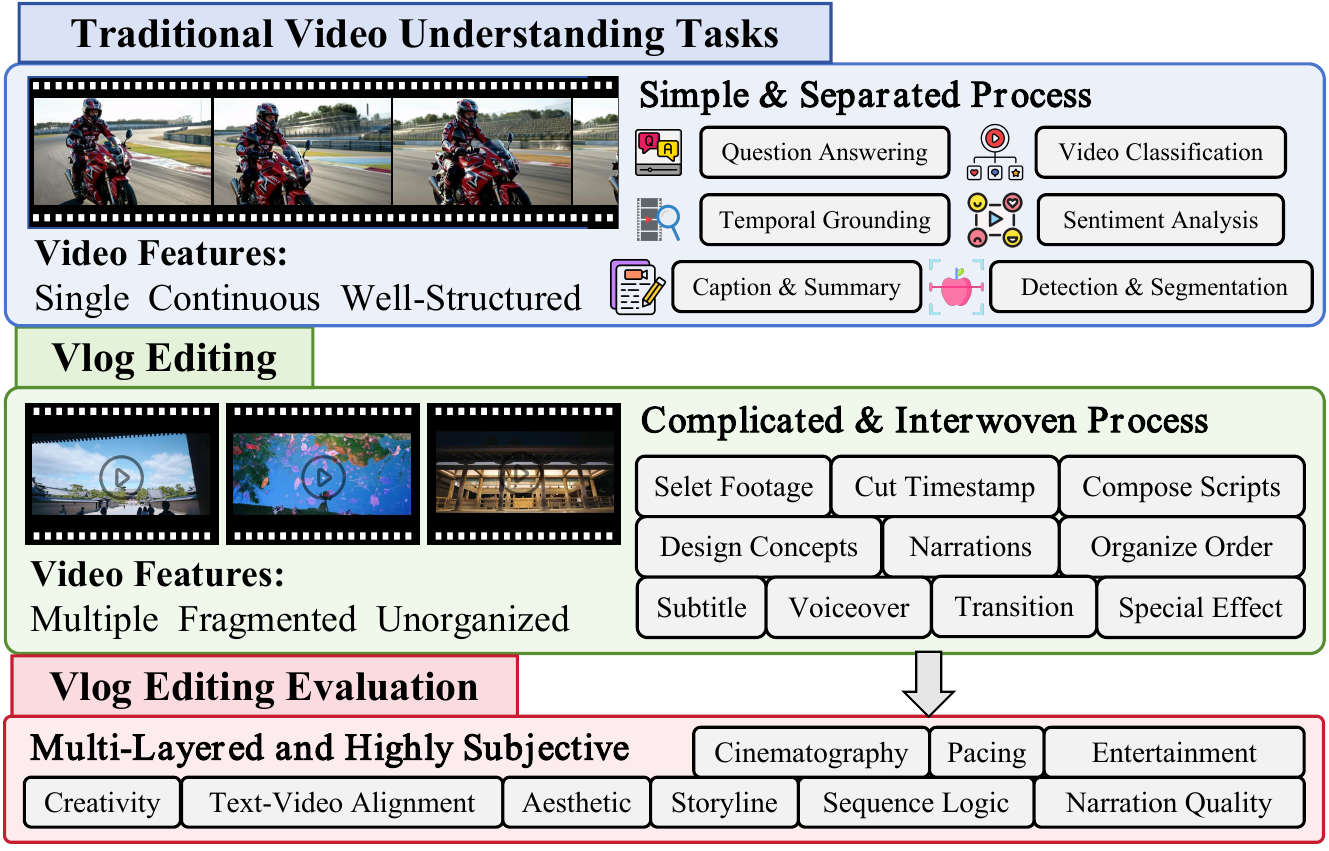}
    \vspace{-2pt}
    \caption{Different from regular video understanding tasks whose processes are often simple and separated, vlog editing invovles more complicated and interwoven processes. To conduct a comprehensive assessment for each step and element of vlog editing, the evaluation criteria would be multi-layered and highly subjective, showcasing unique challenges for vlog reward modeling.}
    \label{Fig1}
    \vspace{-14pt}
\end{figure}

\section{Introduction}

Nowadays, the rapid proliferation of personal media creation has positioned video blogs (\textit{i.e.}, vlogs) as a dominant and highly personalized medium for online storytelling \citep{ladhari2020youtube, duan2025vlog}. Concurrently, the comprehensive capabilities of Multimodal Large Language Models (MLLMs) have developed significantly \citep{zhang2025videollama, hong2025glm, gemini2.5}, offering new possibilities for complex video-language tasks. This synergy raises a new question: \textit{{can we leverage the advanced multimodal reasoning capabilities of MLLMs to evaluate vlog drafts and distill high-quality editing plans?}} This research would help vlog creators improve their artworks, while paving the way for automated vlog edit refinement systems to facilitate commercial production.

Unlike traditional video understanding tasks that typically deal with single, continuous, and well-structured videos \citep{tang2025video,madan2024foundation}, vlog editing is more complicated, as shown in Figure \ref{Fig1}. It starts with a vast collection of raw, unorganized, and often redundant user-provided video assets. To transform these discrete footages into a compelling vlog, creators must navigate a complex decision-making space: designing the storyline and narrative style, selecting the most expressive segments from cluttered assets, precisely cropping timestamps, determining sequence order, composing scripts, and orchestrating multi-modal editing elements such as subtitles, voiceovers, transitions, music, emojis, font style, special effects, etc. 

Consequently, evaluating a vlog's quality is inherently multi-layered and more subjective, presenting unique challenges that transcend standard image/video evaluation \citep{xu2023imagereward, Video-mme}. Beyond objective factors like subtitle-visual correspondence and logic of shot order, the essence of a montage vlog lies in highly subjective dimensions: the concept design of the theme, overall narrative creativity, the cinematographic aesthetic of shot selection, the storyline pacing, etc. These factors are difficult to quantify but essential for distinguishing varied-quality vlog edits. 

Therefore, despite the compelling potential of MLLM-based evaluation \citep{wang2025visualprm,wang2025unified,wu2025rewarddance}, the automatic assessment of vlog edits remains an arduous task. \textbf{First}, there is a critical shortage of standardized vlog evaluation criteria for reference. 
\textbf{Second}, this field suffers from a severe deficit of dedicated benchmarks and annotated vlog data. \textbf{Lastly}, these collectively result in a lack of effective models for such vlog reward modeling.

To systemically overcome these challenges, we \textbf{first} engage professional vlog designers and artistic experts to consolidate the evaluation metrics and define a standardized taxonomy of six key dimensions, establishing a multidimensional and granular assessment framework. \textbf{Second}, we curate a large-scale training dataset with 100k vlog edits, and a dedicated benchmark to evaluate the vlog rewarding capabilities of MLLMs. \textbf{Lastly}, we develop a robust Vlog Reward Model (VRM) that can provide multi-dimensional scores and feedback for any given raw footage assets and corresponding editing plan.

Following recent progress in reward modeling \citep{guo2025reward, wang2025unified2}, we first perform Supervised Fine-Tuning (SFT) and then utilize reinforcement learning (RL) to further enhance our VRM's vlog reward capability.
However, vanilla Group Relative Policy Optimization (GRPO) \citep{shao2024deepseekmath} is inherently designed to calculate intra-group relative advantages, thus lacking cross-sample reference signals for different edits. We observe that this would lead to ``direction blindness'' issue and introduce \textbf{inter-group comparison reward}. By involving reward signals of comparing inter-group rollouts, it enables the model to better distinguish vlog edits of varying quality and more effectively identify the optimal editing plan.

Our contributions can be summarized as follows:
\begin{itemize}
[itemsep=-0.2em, topsep=-0.2em]
    \item \textbf{Novel Task Formulation and Systematic Evaluation Framework.} We propose a new task of automated vlog editing evaluation, and establish a comprehensive taxonomy across six key dimensions in collaboration with professional vlog designers and product managers.

    \item \textbf{Large-Scale Dataset and Benchmark Construction.} We curate a high-quality training dataset of 100k vlog edits, and introduce \textbf{VRMBench}, a dedicated benchmark including both subjective, aesthetic evaluation cases and objective, factual-error ones. VRMBench comprises 400 unique groups, each containing a set of raw video assets paired with four distinct editing plans of descending quality, annotated with scores and textual feedback. These involve a two-stage laborious data collection with ~\textbf{3.6k person-hour} verification.

    \item \textbf{Multi-Dimensional Vlog Reward Model.} We develop \textbf{VlogReward} that can discriminate varied-quality vlog edits with multi-dimensional scores and feedback. With our inter-group comparison reward design, VlogReward achieves superior comparison accuracy (69.4\%$\rightarrow$73.5\%), Best-of-N accuracy (65.0\%$\rightarrow$67.6\%) and score accuracy (50.3\%$\rightarrow$51.3\%). Moreover, adjusting the intensity of comparison reward enables a flexible trade-off between scoring and discriminating capability.
\end{itemize}

\begin{figure*}[!th]
    \centering
    \includegraphics[width=1\linewidth]{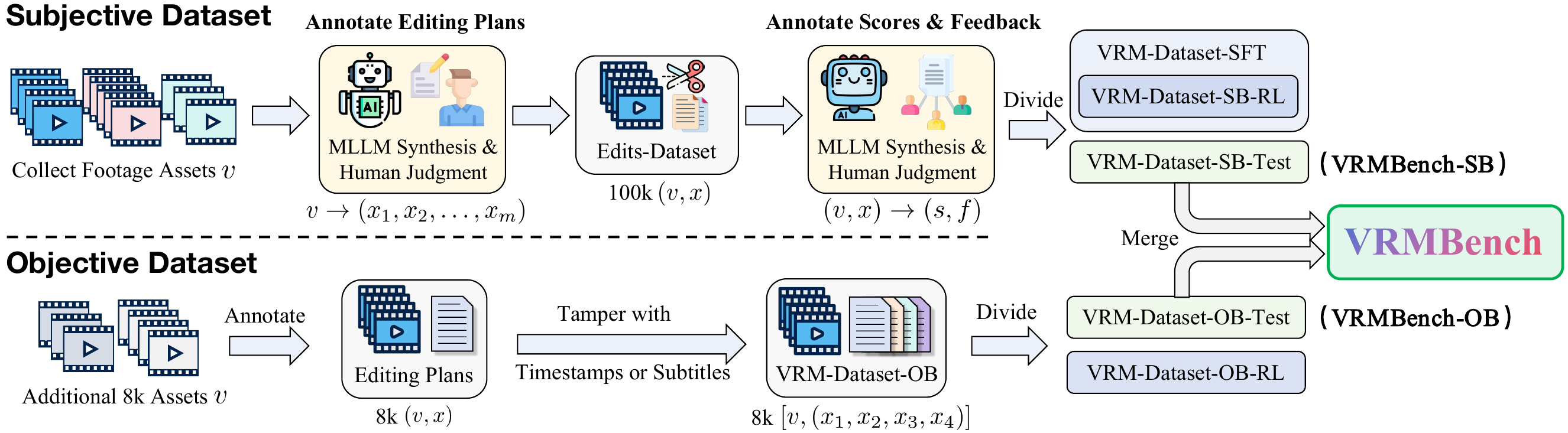}
    \vspace{-0.5pt}
    \caption{Data collection pipeline. Subjective Dataset: We first collect raw assets $\footage$ and use MLLMs to synthesize diverse editing plans $(\edit_1, \edit_2, \dots, \edit_m)$ differing from subjective domains, \textit{e.g.}, aesthetics and creativity. Then we reward them to obtain multi-dimensional scores $\score$ and feedback $\feedback$. The entire process involves human verification to check for hallucinations, ensure alignment between scores and feedback, and filter out low-quality data. Objective Dataset: We additionally collect 8k assets and annotate the editing plans with the same pipeline. Subsequently, we progressively introduce tampering such as altering cut timestamps or subtitles to create objective factual errors, resulting in 4 plans of decreasing quality for each footage. Both the subjective and objective datasets are split separately. The resulting splits are then merged to form the final training dataset and VRMBench benchmark.}
    \label{fig:data collection}
    \vspace{-7pt}
\end{figure*}

\section{Related Work}

\textbf{Vlog Editing and Evaluation.} Due to the complexity of vlog editing, various studies mainly focus on its decomposed sub‑tasks such as highlight detection \citep{gan2023collaborative}, logical ordering of shots \citep{pardo2024generative}, video content summarization \citep{xie2022multimodal}, and next‑shot prediction \citep{hu2023reinforcement}. With the rapid progress of MLLMs, several recent researches have begun to explore agentic end‑to‑end frameworks to automate the entire editing pipeline in a single pass \citep{yang2024agent, sandoval2025editduet,pardo2024generative,wang2025long}. However, existing work widely evaluates vlog editing with IoU‑type overlap metrics or adopts custom, labor‑intensive human rating, which limits reproducibility and promotion. Moreover, vlog editing is an inherently creative and flexible process, and there should be no absolute ground truth label for cut timestamps. This necessitates a standardized, comprehensive, and general evaluation framework, and a corresponding vlog reward model to foster automatic evaluation.

\textbf{Multimodal Reward Modeling.} Multimodal reward modeling plays a pivotal role for aligning MLLMs with human preferences \citep{Mm-rlhf, sun2024aligning, zhang2025basereward,DuanLHHH26}. Early approaches use fine-tuned CLIP models to produce preference scores \citep{kirstain2023pick, wu2023human}. Subsequent works append a linear regression head to the model's backbone to output a scalar reward value \cite{liu2025improving, zang2025internlm}. With the rapid development of MLLMs, generative reward modeling (GRM) represents a paradigm shift from scalar regressive reward models (RRM), enabling more flexible and interpretative output. Recently, several GRM studies have attempted to apply some reasoning strategies, such as Test-Time scaling and RL approaches \citep{liu2025breaking,zhang2026improving,liu2025rethinking} to further enhance the reward capability \citep{Rm-r1, xue2025dancegrpo}. However, most existing multimodal reward modeling frameworks directly utilize a pairwise comparison paradigm to identify relative preferences \citep{zhang2025r1, zhang2025videorewardbench}. While pairwise ranking can efficiently capture comparative signals, it lacks the capacity for fine-grained, multi-dimensional scoring, and straightforward diagnostic feedback. 
Therefore, we propose a GRM framework beyond binary preferences, providing comprehensive scores and feedback across six distinct dimensions while reflecting relative preferences.

\textbf{GRPO Variants.} Several variants of GRPO have emerged to address its stability and scalability challenges. GSPO \citep{zheng2025group} shifts to sequence-level optimization to stabilize Mixture-of-Experts models, and DAPO \citep{yu2025dapo} introduces decoupled clipping for large-scale RL. BNPO \citep{xiao2025bnpo} introduces adaptive reward normalization using a dynamic Beta distribution. GMPO \citep{zhao2025geometric} replaces the arithmetic mean with a geometric mean to suppress token-level outliers. For decentralized environments, GEPO \citep{zhang2025gepo} utilizes group expectation weighting to mitigate high KL divergence caused by network latency. Similar to our inter-group reward idea, some work utilizes inter-group rollouts to calculate advantages. For example, DyKnow-RAG \citep{xu2025dyknow} integrates GRPO to retrieval-augmented generation with inter-group advantage. GRPOformer \citep{guo2025grpoformer} adapts GRPO for efficient hyperparameter optimization using inter-group relative advantages.

\sethlcolor{yellow}
\begin{figure*}[!ht]
    \centering
    \includegraphics[width=0.99\linewidth]{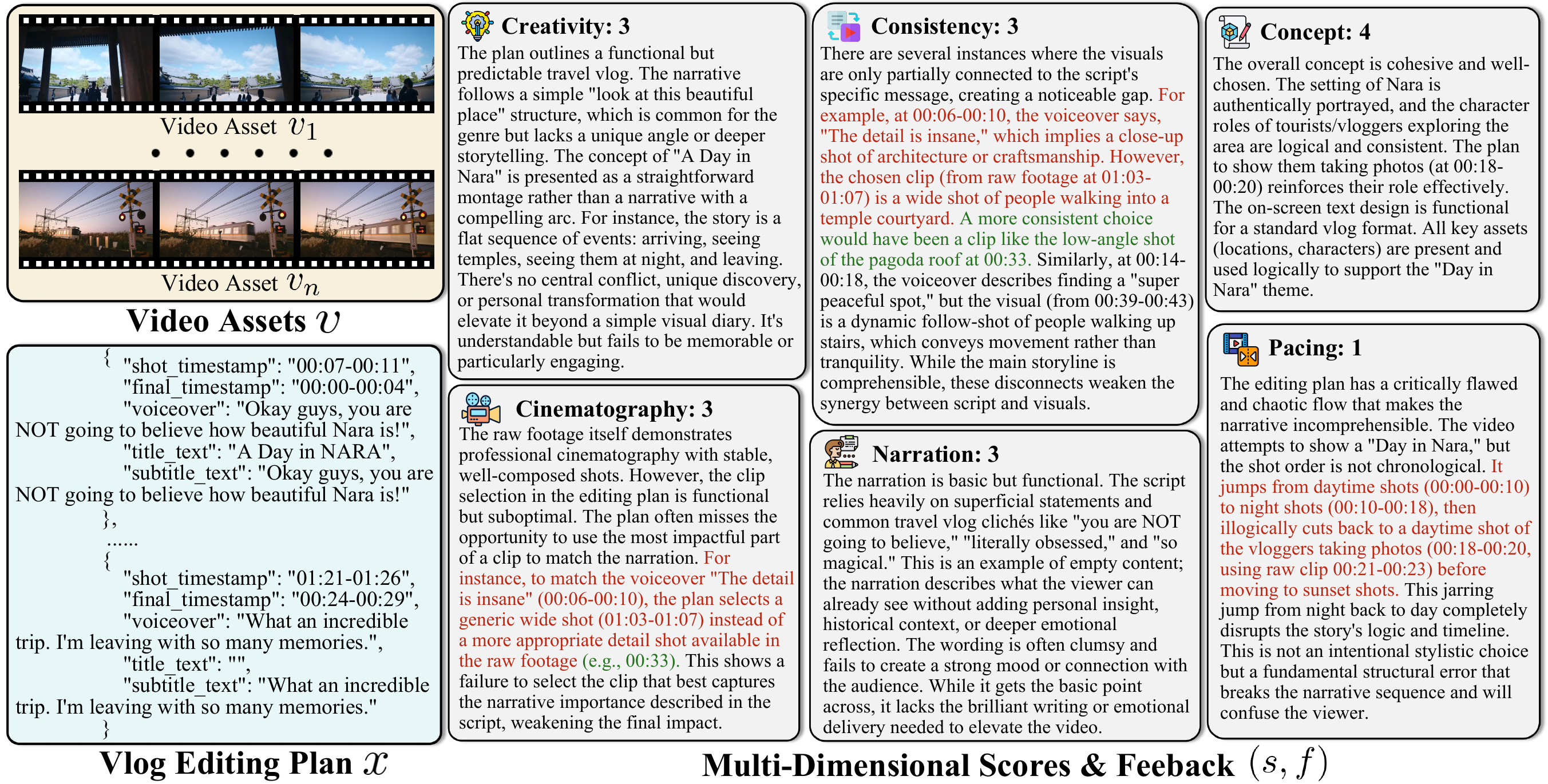}
    \caption{Data example. To facilitate multimodal processing, raw video assets $\{\footage_1, \footage_2, \dots, \footage_n\}$ are concatenated into a single video $\footage$, with each clip separated by a 3-second black screen. The editing plan $\edit$ is a JSON encapsulating the cut timestamps, playback speed, voiceover, shot order, title and subtitle. We restrict editable variables to these core structural attributes to accommodate MLLMs’ frame-sampling mechanism and minimize confounds. Stylistic elements such as voice timbre, transitions, visual effects, and customized typography (\textit{e.g.}, font color, size, and style) are standardized. In this scheme, ``shot\_timestamp'' and ``final\_timestamp'' correspond to the time range in the stitched video $\footage$ and final edited vlog, respectively. A duration mismatch between them triggers a playback speed adjustment. The accompanying feedback provides scoring rationale, where evident, flags \textcolor{MyDarkRed}{notable issues} and \textcolor{MyDarkGreen}{improvement suggestions}.
    }
    \label{fig:data_example}
    \vspace{-8pt}
\end{figure*}

\section{Method}
\subsection{Task Design}
Focusing on Montage vlogs, our core objective is to develop a robust VRM to evaluate the quality of editing plans. Given a set of raw video assets $\footage=\{\footage_1, \footage_2, \dots, \footage_n\}$, the VRM $\model$ should provide a comprehensive assessment of any candidate editing plan $\edit$.

Unlike scalar reward models used in simpler tasks, our VRM is designed as a multi-dimensional generative reward model. For a given pair $(\footage, \edit)$, the model generates a set of numerical scores $\score = \{\score_1, \score_2, \dots, \score_k\}$ representing $k$ distinct dimensions of quality, and an interpretable textual feedback sequence $\feedback = \{\feedback_1, \feedback_2, \dots, \feedback_k\}$. The feedback $\feedback$ serves as a diagnostic critique, explicitly identifying factual issues and providing refinement instructions within the editing plan $\edit$, thereby enhancing the transparency and explainability of the evaluation process, \textit{i.e.}, $\model(\footage, \edit) \rightarrow (\score, \feedback)$.
We use this mechanism as it can be utilized in two ways. First, the numerical score $\score$ can promote to select the optimal editing plan through Test-Time Scaling (TTS) or serve as reward signals to enhance the editing model's performance through RL. Second, the textual feedback $\feedback$ can return to the policy model for iterative refinement of editing plans.

With the guidance of professional vlog creators and product managers, we consolidate the vlog evaluation metrics into six key dimensions, and assign integer ratings from 1 to 5, \textit{i.e.}, $k=6$, $\score_k \in \{1,2,3,4,5\}$. The six evaluation dimensions are \textit{\textbf{Creativity}} (narrative inventiveness and originality), \textit{\textbf{Consistency}} (text-visual alignment), \textit{\textbf{Concept Design}} (script element appropriateness), \textit{\textbf{Cinematography} }(shot selection and framing), \textit{\textbf{Narration} }(storyline and narration quality), and \textit{\textbf{Pacing}} (temporal rhythm and sequence logic). Details of the evaluation criteria are provided in Appendix \ref{appendix criteria}.

\subsection{Dataset}
\label{section dataset}
The dataset collection follows a two-stage pipeline. In the 1st stage, we collect raw footage $\footage$ and corresponding varied-quality editing plans $\{\edit_1, \edit_2, \dots, \edit_m\}$. In the 2nd stage, we reward each pair of $(\footage, \edit), \edit \in \{\edit_1, \edit_2, \dots, \edit_m\}$ to obtain the score and feedback $(\score, \feedback)$. Edits of better overall quality should receive higher total score sums $\sum_{i=1}^k \score_i$, maintaining a consistent quality-to-score mapping. To enhance data diversity, we curate two sets of edits that differ along subjective (SB) or objective (OB) dimensions. Details are expounded in the following and displayed in Figure \ref{fig:data collection}.

\begin{figure}[!ht]
    \centering
    \includegraphics[width=1\linewidth]{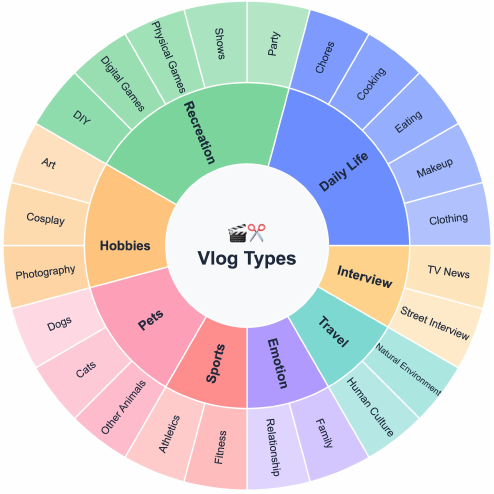}
    \caption{Vlog types. Our dataset contains various vlog types including daily life, recreation, hobbies, pets, sports, emotion, travel, and interview.}
    \label{fig: vlog type}
\end{figure}

\begin{figure*}[!ht]
    \centering
    \includegraphics[width=0.99\linewidth]{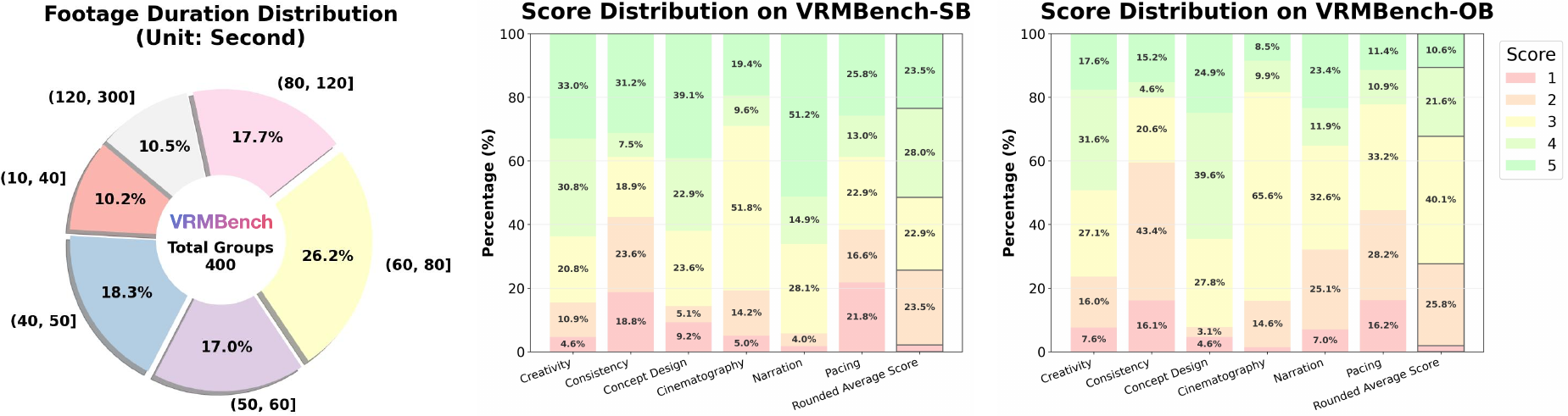}
    \caption{Data Statistics. Our dataset contains various vlog types, including daily life, recreation, hobbies, pets, sports, emotion, travel, and interview. On VRMBench, the footage durations are appropriately distributed, ranging from 10 seconds to 5 minutes. Both VRMBench-SB and VRMBench-OB contain balanced data across all score values, demonstrating the diversity of our dataset.}
    \label{fig:data statistics}
    \vspace{-5pt}
\end{figure*}

\subsubsection{Raw Assets and Editing Plans $(\footage, \edit)$}

We collect extensive raw footage from public platforms spanning diverse categories. Leveraging advanced MLLMs, we identify narrative templates for each group of assets and synthesize the corresponding editing plans. After a rigorous human-in-the-loop verification phase to eliminate low-quality samples, we finally obtain \textbf{Edits-Dataset} comprising raw video assets with standard editing plans. Figure \ref{fig: vlog type} summarizes the various vlog types our dataset contains.

\subsubsection{Scores and Textual Feedback $(\score, \feedback)$}
Given the high subjectivity of vlog evaluation, the annotated scores even vary with the well-defined rubric with the same annotator. To mitigate this issue, we employ an AI-assisted annotation pipeline with human verification. We first design MLLM pipeline to evaluate each $(\footage, \edit)$ from Edits-Dataset to obtain \textbf{VRM-Dataset-SB}.
Then we select samples with 4 edits, where the score difference between any two plans $\geq$ 2. These data are manually audited to ensure that there are no hallucinations and the feedback $\feedback$ is well-aligned with the score $\score$.
Finally, we divided them into two parts: \textbf{VRM-Dataset-SB-RL} (20k) and \textbf{VRM-Dataset-SB-Test} (8k). We exclude VRM-Dataset-SB-Test from VRM-Dataset-SB, and obtain \textbf{VRM-Dataset-SFT} (100k).

To introduce objective criteria, we additionally collect 8k video assets with standard editing plans.
We then randomly select a segment to tamper with shot timestamps or subtitles to create factual errors, ensuring each modification progressively degrades the plan's quality. After automatic scoring, we filter out samples where the human-judged quality ranking contradicted the scores. This results in \textbf{VRM-Dataset-OB} consisting of 5k assets, each paired with 4 editing plans of decreasing quality. We randomly sample 200 assets to obtain \textbf{VRM-Dataset-OB-Test}, remaining others as \textbf{VRM-Dataset-OB-RL}. Finally, VRM-Dataset-SB-RL and VRM-Dataset-OB-RL are merged into \textbf{VRM-Dataset-RL} (40k), and VRM-Dataset-SB-Test and VRM-Dataset-OB-Test are merged into \textbf{VRMBench} (1.6k).

\subsection{Benchmark}
As described in Section \ref{section dataset}, we use VRMBench to test and benchmark the vlog rewarding capability of mainstream MLLMs. VRMBench comprises 400 unique data groups, each consisting of a set of raw footage (3-10 clips) paired with 4 editing plans of descending quality. Each plan is annotated with scores and textual feedback across six dimensions, where the cumulative score strictly follows the decreasing order of quality. Each group features distinct raw assets with total durations ranging from 10 seconds to 5 minutes. The benchmark is categorized into two subsets: VRMBench-SB and VRMBench-OB. VRMBench-SB contains 200 groups where quality variance is driven by subjective attributes (\textit{e.g.}, creativity and aesthetic appeal), while VRMBench-OB includes 200 groups focusing on objective discrepancies (\textit{e.g.}, factual errors in subtitle descriptions). More statistics are displayed in Figure \ref{fig:data statistics}.

\subsection{GRPO with Inter-Group Comparison Reward}
\label{section method grpo}
\begin{figure}[t]
  \centering
  \includegraphics[width=0.46\textwidth]{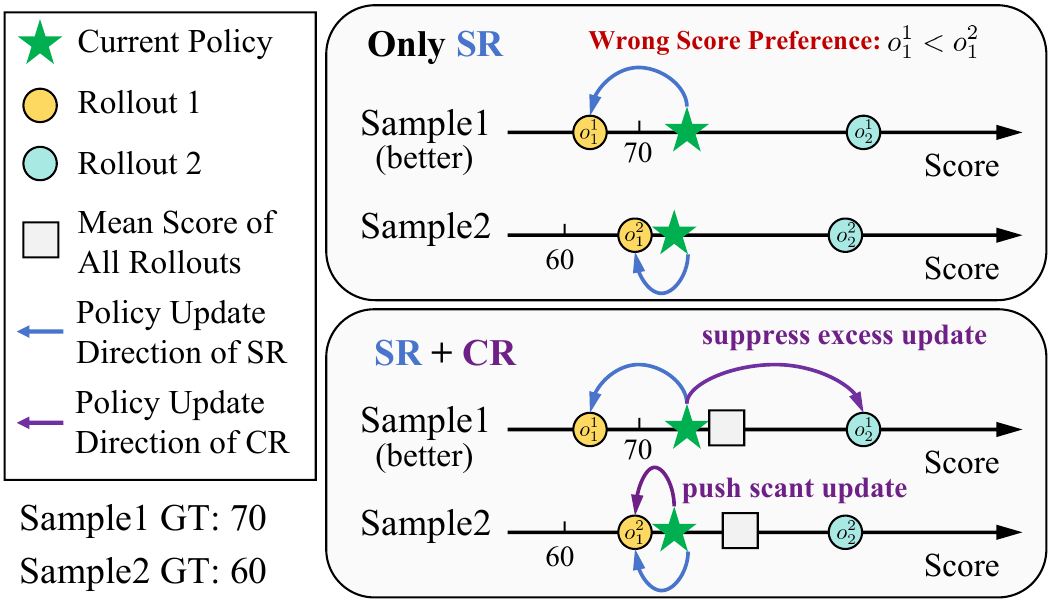}
  \caption{Illustration of the role of comparison reward (CR) beyond score reward (SR). For ease of analysis, we consider a simplified scenario with 2 rollouts and only score in a single dimension from 0 to 100. We assume the vlog edit sample 1 is better than sample 2, with ground truth (GT) score 60 and 70, respectively. For both samples, the predicted score of rollout 1 is closer to GT, resulting in higher SR and positive relative advantage. Consequently, the policy update direction favors rollout 1 and approaches GT. However, this would lead to lower predicted scores of sample 1 compared to sample 2, \textit{i.e.}, $o^1_{1} < o^2_{1}$ for their scores. In contrast, CR adds a direction signal of score difference between sample 1 and 2 that SR alone lacks. The limitation of SR rises to this ``direction blindness'' issue, failing to recognize the preference direction. With SR + CR, $o^1_{2}$ exhibits higher CR and provides an opposing policy update direction to restrain excessive updates of SR. By appropriately adjusting the intensity of CR, we can maintain score accuracy while ensuring score differences align with sample quality preferences.}
  \vspace{-28pt}
  \label{fig rollout}
\end{figure}

We use the pipeline of SFT + GRPO to train our VRM. However, GRPO inherently computes rewards and relative advantages within intra-group rollouts and lacks inter-group comparative information. Despite improving the accuracy of absolute scores, it suffers from ``direction blindness'' issue (explained in Figure \ref{fig rollout}). 
One potential remedy is to calculate inter-group advantages in a batch \citep{guo2025grpoformer}. However, the high variance between different samples may lead to training instability, where a positive-advantage rollout for a single sample may be improperly penalized with negative advantage relative to inter-group rollouts. 
To address this, we introduce inter-group comparison reward while ensuring relative advantages are still computed within intra-group rollouts of the single sample. 
Given a sample of raw assets and an editing plan $q=(\footage, \edit)$, the old VRM policy model $\pi_{\mathrm{old}}$ generates a group of rollouts $\{o_1, o_2, \dots, o_G\}$. We optimize the current policy model $\pi_\theta$ by maximizing the objective of standard GRPO, as described in Appendix \ref{appendix implementation}. Our reward designs are elaborated in the following. 

\textbf{Format Reward.} We do not use the common thinking mode that encloses the thought process with \texttt{<think>...</think>} and the answer with \texttt{<answer>...</answer>}, as it yeilds sub-optimal performance compared to only using the answer content. Details can be found in Appendix \ref{appendix thinking mode}. The VRM is required to provide integer scores from 1 to 5 and feedback for all six dimensions in its output answer.

\begin{equation}
    \reward_{\mathrm{format}} = 
\begin{cases} 
0, & \text{if output matches format}, \\
1, & \text{if output doesn't match format}.
\end{cases}
\end{equation}

\textbf{Score Reward (SR).} The score reward is formulated as the proportion of the six evaluation dimensions that match the ground-truth labels. Specifically, let $\score = \{\score_1, \score_2, \dots, \score_k\}$ be the predicted scores and $\mathbf{\hat{s}} = \{\hat{\score}_1, \hat{\score}_2, \dots, \hat{\score}_k\}$ be the corresponding ground truths across $k=6$ dimensions. We assign a nominal reward $\alpha$ (~\ie 0.1 in our setting) to rollouts that maintain the correct format but fail to match any ground-truth scores. This serves to distinguish such instances from malformed outputs, which are penalized with a reward of 0. The score reward is defined as
\begin{equation}
    \reward_{score} = \frac{1}{k} \sum_{i=1}^{k} \mathds{1}(\score_i=\hat{\score_i}) \times (1-\alpha) + \alpha.
\end{equation}

\textbf{Comparison Reward (CR).} 
To further enhance the VRM's ability to differentiate varied-quality editing plans, we design a comparison reward to introduce comparative information among inter-group rollouts. In VRM-Dataset-RL, each raw assets $\footage$ is paired with $m=4$ editing plans $\{\edit^1, \edit^2, \edit^3, \edit^4\}$ of descending quality. Note the sum of predicted scores for $(\footage, \edit^j)$ as $\scoresum^j = \sum_{i=1}^{k} \score^j_i$ and the ground-truth sum as $\hat{\scoresum}^j = \sum_{i=1}^{k} \hat{\score}_i^j$, it satisfies $\hat{\scoresum}^1 > \hat{\scoresum}^2 > \hat{\scoresum}^3 > \hat{\scoresum}^4$. The core idea is to encourage VRM to assign higher predicted scores to better samples $(\footage, \edit^j)$ than worse samples $\{(\footage, \edit^{l})\mid j>l\}$, \textit{i.e.}, $\scoresum^j > \scoresum^{l}$. The comparison reward for the $i$th rollout $o_i^j$ of $(\footage, \edit^j)$ is defined as
\begin{equation}
    \!\reward_{\mathrm{comp}}(o_i^j,l)=\begin{cases}
        1, \text{if} \,\,\text{sgn}\big[(\scoresum_i^j - \frac{\sum_{i'=1}^{G'}\scoresum_{i'}^l}{G'} )(\hat{\scoresum}^j - \hat{\scoresum}^l)\big]=1, \\
        0,\, \text{otherwise}.
    \end{cases}
\end{equation}
\begin{equation}
    \reward_{\mathrm{comp}}(o_i^j) = \frac{1}{m-1} \sum_{\substack{l=1 \\ l \neq j}}^m \reward_{comp}(o^j_i, l) ,
\end{equation}
where $\scoresum^j_i$ is the sum of predicted scores of $o_i^j$, $m=4$ is the number of candidate plans, and $G'$ is the number of rollouts in the correct format.

\textbf{Total Reward.} The total reward is defined as
\begin{equation}
    \reward_{\mathrm{total}} = \reward_{\mathrm{format}} \times \big[(1 - \lambda) \cdot \reward_{\mathrm{score}} + \lambda \cdot\reward_{\mathrm{comp}}\big],
\label{reward_total}
\end{equation}
where $\lambda$ is the hyper-parameter to adjust the intensity of the comparative reward signal.

\begin{table*}[ht]
    \centering
    \caption{Comprehensive evaluations of different MLLMs and baselines on VRMBench. SB refers to VRMBench-SB, OB represents VRMBench-OB, and AVG indicates the average performance between SB and OB. Compared with other SOTA MLLMs and different VRM counterparts, our model VlogReward achieves the highest average performance on all metrics.}
    \label{tab1}
    \setlength{\tabcolsep}{8.5pt}
    \renewcommand{\arraystretch}{0.87} 
    \begin{tabular}{l|ccc|ccc|ccc}
        \toprule
        \multirow{2}{*}{\textbf{Models}} & \multicolumn{3}{c|}{\textbf{Score Accuracy}} & \multicolumn{3}{c|}{\textbf{Comparison Accuracy}} & \multicolumn{3}{c}{\textbf{Best-of-N Accuracy}} \\
        \cmidrule{2-10}
        & SB & OB & AVG & SB & OB & AVG & SB & OB & AVG \\
        \midrule
        \tabtitlecolor
        \multicolumn{10}{c}{\textit{Closed-Source MLLMs}} \\ [-2pt]
        \midrule
        GPT-4o & 24.4 & 32.4 & 28.4 & 38.1 & 51.8 & 44.9 & 24.3 & 42.1 & 33.2 \\
        GPT-5 & 29.7 & 43.2 & 36.4 & 53.7 & 72.5 & 63.1 & 37.4 & 65.7 & 51.5 \\
        o4-mini & 24.3 & 38.2 & 31.3 & 46.1 & 64.6 & 55.3 & 26.5 & 47.7 & 37.1 \\
        Gemini-3-Pro & 35.0 & 42.9 & 38.6 &\textbf{60.5} & 83.3 & \underline{71.9} & 35.5 & 70.1 & 52.8 \\
        \midrule
        \tabtitlecolor
        \multicolumn{10}{c}{\textit{Open-Source MLLMs}} \\ [-2pt]
        \midrule
        LLaVA-OneVision-7B & 24.0 & 26.7 & 25.4 & 40.9 & 44.8 & 42.8 & 18.9 & 21.9 & 20.4 \\
        LLaVA-NeXT-Video-7B & 25.5 & 31.5 & 28.5 & 39.8 & 44.7 & 42.2 & 24.4 & 28.0 & 26.2 \\
        InternVL3.5-14B & 23.2 & 29.5 & 26.3 & 30.2 & 36.8 & 33.5 & 20.3 & 35.1 & 27.7 \\
        Kimi-VL-A3B-Thinking-2506 & 22.9 & 30.4 & 26.7 & 41.6 & 44.8 & 43.2 & 22.7 & 23.6 & 23.1 \\
        MiMo-VL-7B-RL & 22.5 & 31.4 & 26.9 & 44.4 & 53.9 & 49.2 & 28.5 & 39.1 & 33.8 \\
        Qwen2.5-VL-7B & 22.4 & 32.9 & 27.7 & 34.8 & 42.8 & 38.8 & 20.5 & 31.9 & 26.2 \\
        Qwen2.5-VL-72B & 26.1 & 38.1 & 32.1 & 36.8 & 55.3 & 46.0 & 24.8 & 44.5 & 34.6 \\
        Qwen3-VL-8B & 22.8 & 31.6 & 27.2 & 39.3 & 54.8 & 47.0 & 24.9 & 47.0 & 35.9 \\
        \midrule
        \tabtitlecolor
        \multicolumn{10}{c}{\textit{Regressive Vlog Reward Models}} \\ [-2pt]
        \midrule
        RRM-MSE & 23.6 & 25.4 & 24.5 & 42.3 & 45.5 & 43.9 & 31.6 & 41.0 & 36.3 \\
        RRM-BT & - & - & - & 58.4 & 83.1 & 70.8 & 36.5 & 69.5 & 53.0 \\
        \midrule
        \tabtitlecolor
        \multicolumn{10}{c}{\textit{Generative Vlog Reward Models}} \\ [-2pt]
        \midrule
        GRM-SFT & 37.4 & 41.4 & 39.4 & 54.7 & 65.0 & 59.8 & 40.3 & 44.1 & 42.2 \\
        GRM-GRPO & \underline{44.0} & \underline{56.5} & \underline{50.3} & 54.8 & \underline{84.0} & 69.4 & \underline{43.7} & \underline{86.2} & \underline{65.0} \\
        GRM-GRPO-Inter-Advantage & 43.6 & 54.6 & 49.1 & 38.8 & 73.3 & 56.0 & 35.7 & 75.6 & 55.7 \\
        \textbf{VlogReward (Ours)} & \textbf{45.9} & \textbf{56.7} & \textbf{51.3} & \underline{60.4} & \textbf{86.6} & \textbf{73.5} &\textbf{ 48.3} & \textbf{86.8} & \textbf{67.6} \\
        \midrule
    \end{tabular}
    \vspace{-5pt}
\end{table*}

\section{Experiments}
\subsection{Settings}
\textbf{Dataset and Benchmark.} 
As described in Section \ref{section dataset}, VRM-Dataset-SFT (100k) is employed as SFT training data for cold start initialization. After that, we use the refined dataset VRM-Dataset-RL (40k) to perform RL. VRMBench is used to test and benchmark other MLLMs. 

\textbf{Implementation Details.} We use Qwen2.5-VL-7B-Instruct \citep{bai2025qwen2.5} as our base model. For the trade-off between performance and efficiency, we limit the fps to 1 and the maximum video frames to 64. Each frame is processed at a maximum resolution of $64 \times 28 \times 28$ pixels.

For all evaluations, we follow the decoding configuration used in the official Qwen2.5-VL demo, with top\_p = 0.001 and temperature = 0.01. 
For SFT training, the learning rate is set to 2e-6, and the total batchsize is 128. After 2 epochs of SFT training on VRM-Dataset-SFT, we conduct RL on VRM-Dataset-RL for 1 epoch, with learning rate 1e-6 and batchsize 64. The rollout temperature is set to 1, and the rollout number $G$ is 8. The hype-parameter $\lambda$ is set to 0.25.

\textbf{Baselines.} The evaluations are compared with recent state-of-the-art MLLMs, including GPT-4o-2024-05-13 \citep{gpt-4o}, GPT-5-2025-08-07 \citep{gpt-5}, o4-mini-2025-04-16 \citep{o4-mini}, Gemini-3-Pro \citep{Gemini-3}, LLaVA-OneVision \citep{Llava-onevision}, LLaVA-Next-Video \citep{zhang2024llavanextvideo}, InternVL3.5 \citep{wang2025internvl3_5}, Kimi-VL-Thinking \citep{kimiteam2025kimivltechnicalreport}, MiMo-VL-RL \citep{coreteam2025mimovltechnicalreport}, Qwen2.5-VL \citep{bai2025qwen2.5} and Qwen3-VL \citep{qwen3technicalreport}. Inference parameters are consistent as described above across all models, except in two cases: 1) GPT-5 and o4-mini use temperature=1 and top\_p=0.7 for official mandatory specifications, and 2) LLaVA-One-Vision and LLaVA-Next-Video use a fixed 16 frames because of their context length constraint. What's more, We also compare different paradigms of our VRM:
\begin{itemize}[itemsep=-0.2em, topsep=-0.2em]
    \setlength{\itemindent}{-1.1em}
    \item \textbf{RRM-MSE}: Replace the LLM head with a trainable linear head to predict scores with mean square error loss.
    \item \textbf{RRM-BT}: Replace the LLM head with a trainable linear head to predict pairwise preferences with Bradley-Terry loss \citep{bradley1952rank}.
    
    \item \textbf{GRM-SFT:} The base MLLM is cold-start initialized with VRM-Dataset-SFT, resulting in this model.
    \item \textbf{GRM-GRPO:} This represents the baseline approach that applies GRPO with only SR on GRM-SFT.
    \item \textbf{GRM-GRPO-Inter-Advantage:} This calculates the relative advantage in inter-group rollouts with the same settings of GRM-GRPO.
    \item \textbf{VlogReward:} This is our final vlog reward model, which performs GRPO with both SR and CR.
\end{itemize}

\subsection{Benchmark Metrics}
\begin{itemize}[itemsep=-0.2em, topsep=-0.2em]
    \item \textbf{Score Accuracy:} The average accuracy of predicted scores on the benchmark $\dataset$, \textit{i.e.}, $\mathbb{E}_{(\footage, \edit) \sim \dataset} \, \frac{1}{k} \sum_{i=1}^k \mathds{1}(\score_i = \hat{\score}_i)$.
    \item \textbf{Comparison Accuracy:} The proportion of assigning higher total score to superior editing plans over inferior ones, \textit{i.e.}, $\mathbb{E}_{(\footage, \edit_j, \edit_l) \sim \dataset, \, j > l} \, \mathds{1}(\scoresum^j > \scoresum^l)$.
    \item \textbf{Best-of-N Accuracy:} The proportion of assigning the highest total score to the best editing plan, \textit{i.e.}, $\mathbb{E}_{(\footage, \edit_1, \edit_2, \dots, \edit_m) \sim \dataset} \, \mathds{1}[(\operatorname*{argmax}\limits_{j} \, \scoresum^j)=1; 1 \leq j \leq m]$. In cases of tied score sums, we randomly select one of the tied plans as the predicted best candidate. This repeats 5 times to report average Best-of-N Accuracy.
\end{itemize}

\begin{figure*}
    \centering
    \includegraphics[width=0.95\linewidth]{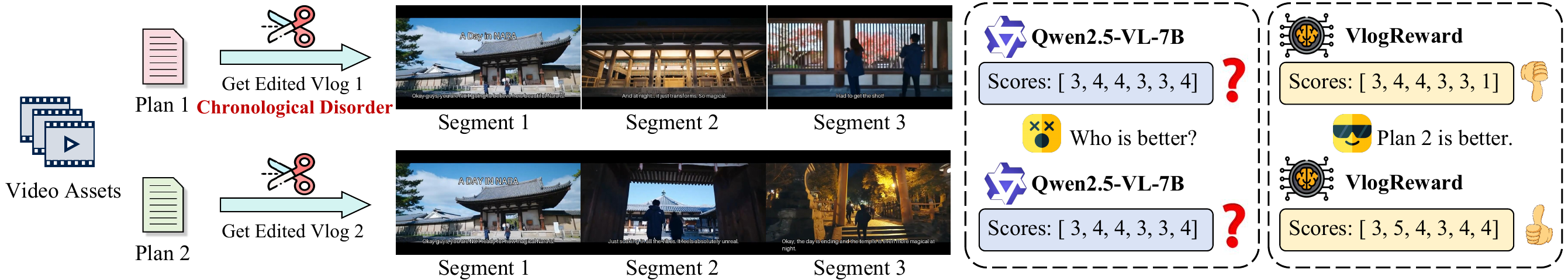}
    \caption{Comparison of reward scores of Qwen2.5-VL-7B and VlogReward for different editing plans. Plan 1 suffers from chronological disorder (day $\rightarrow$ night $\rightarrow$ day), while Plan 2 maintains a coherent temporal flow. Qwen assigns identical scores to both plans, failing to distinguish the edit quality. Conversely, VlogReward accurately identifies the sequence logic flaws in Plan 1 (assigning \textit{pacing} score to 1), and the score difference reflects the correct preference.}
    \label{fig:qwen_example}
    \vspace{-8pt}
\end{figure*}

\subsection{Main Results}
Table \ref{tab1} summarizes the detailed performance of our VRM against various state-of-the-art (SOTA) closed-source and open-source MLLMs. Our analysis yields several key insights into the current state of multimodal vlog evaluation.

\textbf{General Vlog Editing Evaluation Limitations of Existing MLLMs.} The experiment results reveal that the general performance for vlog editing evaluation of existing MLLMs remains deficient. Even when disregarding potential scoring bias, most models exhibit a limited capacity to distinguish varied-quality editing plans, as evidenced by the relatively low Comparison Accuracy and Best-of-N Accuracy across the board. For instance, all tested open-source models fail to surpass 50\% in average Comparison Accuracy, suggesting that current MLLMs struggle to identify the better editing plan among candidates, \textit{i.e.}, assigning equal or even lower scores for better vlog edits. These findings underscore a significant gap in the comprehension capabilities of current MLLMs for vlog rewarding tasks, indicating substantial room for future improvement.

\textbf{Performance Disparity on Objective vs. Subjective Domains.} 
A consistent trend across all evaluated models is that performance on VRMBench-OB is generally superior to that on VRMBench-SB. Existing MLLMs demonstrate higher accuracy in identifying factual errors, but struggle more with subjective criteria about aesthetic feelings.
For example, GPT-5 achieves a 72.5\% Comparison Accuracy on objective dimensions, while dropping to 53.7\% when evaluating nuanced, aesthetic preferences. This disparity highlights the inherent difficulty in modeling complex, human-centric quality metrics compared to explicit factual inconsistencies.

\textbf{Closed-Source Performance Dominance.} Closed-source MLLMs generally outperform open-source counterparts across most metrics. Among the closed-source MLLMs, Gemini-3-Pro emerges as the strongest baselines. It demonstrates superior Comparison Accuracy of 71.9\% and Best-of-N Accuracy of 52.8\%. In contrast, open-source MLLMs show a noticeable performance lag. 

\textbf{Superiority of Our Proposed VlogReward and CR.} VlogReward achieves state-of-the-art results across almost all evaluated metrics, significantly outperforming both open-source and closed-source models, and other VRM counterparts. Figure \ref{fig:qwen_example} shows an example of predicted scores of Qwen2.5-VL-7B and VlogReward. Qwen assigns identical scores to two clearly differentiated editing plans of varying quality, while VlogReward provides targeted scores reflecting editing issues. These demonstrate the prominent ability of our model and our comparison reward design to provide multi-demenisonal scores and identify the best editing plan.

\begin{table}[t]
    \centering
    \small  
    \caption{Detailed performances across different $\lambda$ values.}
    \label{tab2}
    \setlength{\tabcolsep}{5.4pt} 
    \renewcommand{\arraystretch}{0.96}
    \begin{tabular}{l|c| ccccc}
        \toprule
        \multirow{2}{*}{\textbf{Metric}} & \multirow{2}{*}{\textbf{Dataset}} & \multicolumn{5}{c}{\textbf{$\lambda$}} \\
         \cmidrule{3-7}
        & & 0 & 0.25 & 0.5 & 0.75 & 1 \\
        \midrule
        
        \multirow{3}{*}{\makecell[l]{\textbf{Score}\\\textbf{Accuracy}}} & SB & 44.0 & \textbf{45.9} & \underline{44.1} & 43.4 & 21.7 \\
        & OB & \underline{56.5} & \textbf{56.7} & 56.1 & 54.2 & 18.9 \\
        & AVG & \underline{50.3} & \textbf{51.3} & 50.1 & 48.8 & 20.3 \\
        \midrule
        
        \multirow{3}{*}{\makecell[l]{\textbf{Comparison}\\\textbf{Accuracy}}} & SB & 54.8 & 60.4 & 61.2 & \textbf{65.9} & \underline{65.6} \\
        & OB & 84.0 & 86.6 & 88.8 & \underline{90.0} & \textbf{90.2} \\
        & AVG & 69.4 & 73.5 & 75.0 & \textbf{78.0} & \underline{77.9} \\
        \midrule
        
        \multirow{3}{*}{\makecell[l]{\textbf{Best-of-N}\\\textbf{Accuracy}}} & SB & 43.7 & 48.3 & 48.5 & \textbf{49.4} & \underline{49.3} \\
        & OB & 86.2 & 86.8 & 93.4 & \underline{93.2} & \textbf{93.5} \\
        & AVG & 65.0 & 67.6 & 71.0 & \underline{71.3} & \textbf{71.4} \\
        \bottomrule
    \end{tabular}
    \vspace{-8pt}
\end{table}

\begin{table}[htbp]
\centering
\small
\caption{Sensitivity analysis of the Comparison Reward (CR) under different configurations of group size $m$ and rollout number $G$.}
\label{tab:sensitivity_results}
    \setlength{\tabcolsep}{5.4pt} 
    \renewcommand{\arraystretch}{0.96}
\begin{tabular}{cccccc}
\toprule
\boldmath{$m$} & \boldmath{$G$} & \textbf{Reward} & \textbf{Score} & \textbf{Comparison} & \textbf{BoN} \\
\midrule
3 & 8 & SR & 49.1 & 64.8 & 64.3 \\
3 & 8 & \textbf{SR+CR} & \textbf{50.2} & \textbf{68.8} & \textbf{67.0} \\
\midrule
4 & 4 & SR & \textbf{51.0} & 67.9 & 65.3 \\
4 & 4 & \textbf{SR+CR} & 50.2 & \textbf{72.0} & \textbf{66.8} \\
\bottomrule
\end{tabular}
 \vspace{-10pt}
\end{table}

\subsection{Analysis and Discussions}
\begin{figure}[!t]
    \centering
    \includegraphics[width=0.91\linewidth]{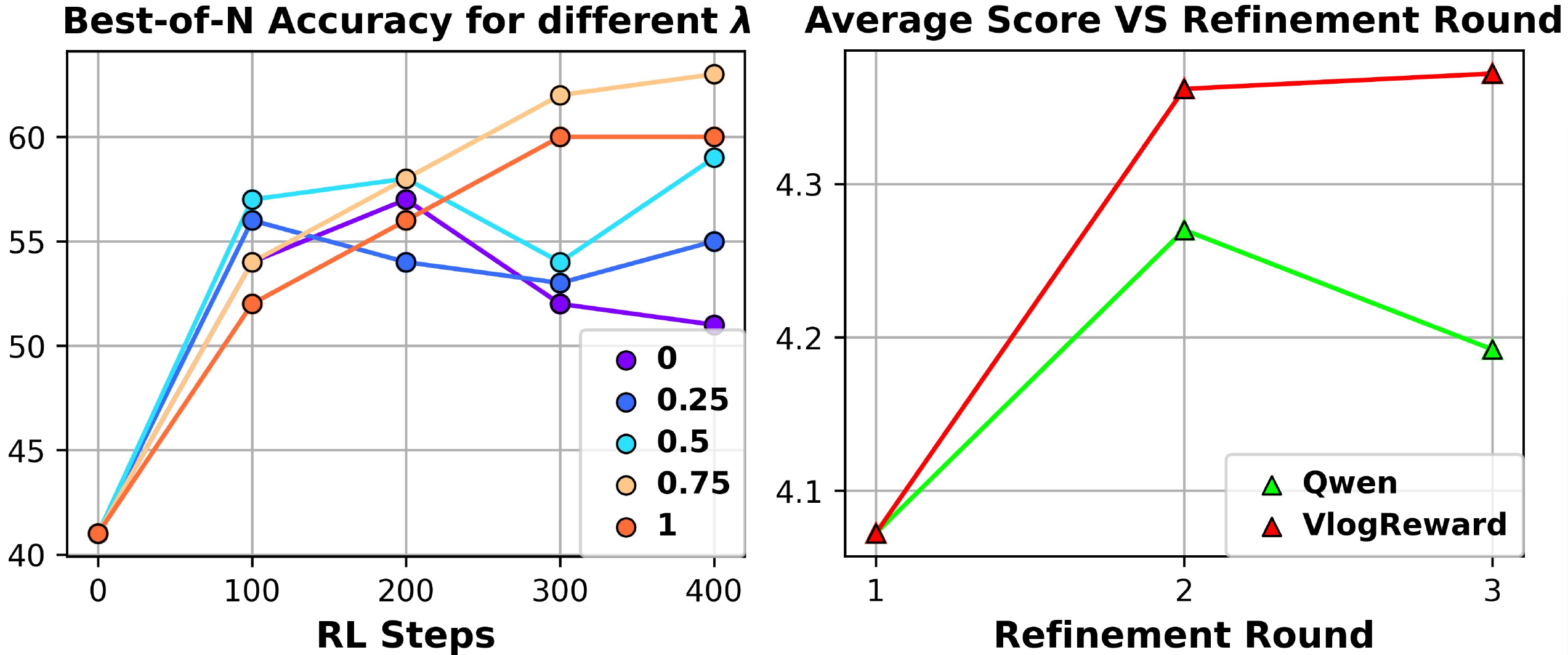}
    \caption{\textit{Left:} Best-of-N Accuracy curve for RL with different $\lambda$ on random 100 samples from VRMBench. \textit{Right:} Average score on all dimensions and samples vs refinement round.}
    \label{fig:refinement}
    \vspace{-15pt}
\end{figure}

\textbf{Effect of Different Values of $\lambda$.} We investigate the influence of the hyper-parameter $\lambda$ in the total reward formulation (Equation \ref{reward_total}), which balances the granular score reward $\reward_{score}$ and the inter-group comparison reward $\reward_{comp}$. As summarized in Table \ref{tab2}, the model's performance exhibits a non-monotonic trend on Score Accuracy as $\lambda$ increases from 0 to 1. Specifically, Score Accuracy initially improves and then gradually declines, reaching its peak performance of 51.3\% at $\lambda=0.25$. Both Comparison Accuracy and Best-of-N Accuracy ascend with the enhancement of the comparison signal, achieving their respective optima of 78\% and 71.4\% at $\lambda=0.75$ and $\lambda=1$, respectively. These indicate that the appropriate introduction of comparison reward can simultaneously enhance both scoring accuracy and the capability to distinguish between superior and inferior samples (\textit{e.g.}, $\lambda=0.25$). The hyper-parameter $\lambda$ serves as an effective mechanism for balancing performance trade-offs across different evaluation metrics. What's more, Figure \ref{fig:refinement} shows the Best-of-N Accuracy curve of RL training. The performance for $\lambda=0.75,1$ shows continuous improvement. However, it exhibits an initial increase followed by a decline for $\lambda=0$, which is most likely attributable to the ``direction blindness'' issue inherent to the score reward.

\textbf{Impact of Group Size $m$.} We evaluate the performance when the group size is reduced to $m=3$ while keeping $G=8$. As shown in Table~\ref{tab:sensitivity_results}, although the reduction in group size slightly limits the comparison space, the integration of CR (SR+CR) still yields substantial improvements over the baseline SR. Specifically, the Comparison accuracy and BoN accuracy increase from 64.8\% and 64.3\% to 68.8\% and 67.0\%, respectively, demonstrating that CR remains highly effective even with fewer candidate plans.

\textbf{Sentivity on Rollout Number $G$.} We restrict the rollout number to $G=4$ while maintaining $m=4$. Under this resource-constrained setting, the baseline SR achieves a score of 51.0, but its comparison and BoN accuracies are limited to 67.9\% and 65.3\%. When applying our CR, we observe a significant boost in comparison accuracy (72.0\%) and BoN accuracy (66.8\%). We note a minor decrease in the absolute reward score (from 51.0 to 50.2), which we attribute to the fact that fewer rollouts ($G=4$) lead to a less stable estimation of the mean baseline in GRPO, thereby slightly perturbing the absolute score scale. Nevertheless, the consistent gains in comparison and BoN selection highlight the robust preference-distinguishing capability of CR.

\textbf{Utilization for TTS Vlog Editing Inference.} Our VlogReward can also be used to enchance the vlog editing performance through Test-Time Scaling. To prove this, we use Edits-Dataset to train a Vlog Editing Model (VEM) to generate vlog editing plans for given raw footage. For evaluation, we randomly sample 100 assets from the VRMBench. The VEM is used to generate $N=8$ candidate editing plans per asset using a sampling temperature of 0.5. We then employ VlogReward to score these candidates and select the plan with the highest total score. We compare this ``Best-of-8'' selection against a baseline generated via greedy decoding. Our user study indicates that the TTS approach with VlogReward scoring was preferred in 62\% of cases, demonstrating a significant improvement in editing quality over standard greedy decoding.

\textbf{Iterative Refinement with Textual Feedback.} To evaluate the quality of the textual feedback provided by VlogReward, we conducte an experiment using the same 100 samples. Specifically, we compare the feedback capabilities of VlogReward against the base model. In each iteration, the VRM provides diagnostic critiques, which the VEM then uses to refine the previous editing plan. As illustrated in Figure \ref{fig:refinement}, the average scores of the editing plans refined by VlogReward show a consistent upward trend and achieve higher overall quality. In contrast, the refinement process guided solely by the base model's feedback results in unstable score trajectories and lower performance. 
These results demonstrate that our VRM provides more accurate and actionable feedback, effectively driving the iterative improvement.

\subsection{Transferability to General Preference Learning}
\label{subsec:transferability}

To evaluate the generalization capability of our proposed CR beyond the domain of vlog editing, we apply it to a broader reward preference learning task within the Reinforcement Learning from Human Feedback framework. 

\textbf{Experimental Setup.} We adopt Qwen2.5-7B-Instruct as our backbone model. For the training and evaluation, we utilize the \texttt{UltraFeedback} dataset \citep{cui2024ultrafeedback}, a large-scale, high-quality benchmark designed for training robust reward models. Each sample comprises a prompt, four model-generated responses, and corresponding fine-grained human/GPT-4 feedback scored across four dimensions: instruction-following, truthfulness, honesty, and helpfulness. In our setup, we allocate 40k samples for SFT, 10k samples for GRPO, and reserve 1k samples for evaluation.

\textbf{Results and Discussion.} Table~\ref{tab:transfer_results} summarizes the evaluation reults on raw reward score accuracy, pairwise comparison accuracy, and Best-of-N (BoN) selection accuracy. Upon integrating SR+CR, the performance significantly improves across all metrics, achieving 47.0 in score, 56.4\% in comparison accuracy, and 53.6\% in BoN accuracy. These results demonstrate that our CR mechanism is not limited to vlog editing but generalizes effectively to general preference alignment tasks, successfully mitigating direction blindness in broader policy optimization scenarios.

\begin{table}[!t]
\small
\centering
\caption{Generalization results on the \texttt{UltraFeedback} dataset.}
\vspace{-2pt}
\setlength{\tabcolsep}{5pt} 
    \renewcommand{\arraystretch}{0.96}
\label{tab:transfer_results}
\begin{tabular}{lcccc}
\toprule
\textbf{Method} & \textbf{Reward} & \textbf{Score} & \textbf{Comparison} & \textbf{BoN} \\
\midrule
SFT & — & 43.3 & 45.9 & 44.0 \\
GRPO & SR & 46.1 & 54.7 & 51.2 \\
GRPO & \textbf{SR+CR} & \textbf{47.0} & \textbf{56.4} & \textbf{53.6} \\
\bottomrule
\end{tabular}
\vspace{-18pt}
\end{table}

\section{Conclusion}

In this paper, we establish a comprehensive six-dimensional taxonomy for vlog editing evaluation, curate a large-scale training dataset and VRMBench benchmark, and develop a robust Vlog Reward Model that provides both fine-grained scores and diagnostic feedback. Our proposed inter-group comparison reward effectively alleviates the "direction blindness" issue of standard GRPO, enabling superior performance in distinguishing varied-quality edits.







\newpage
\section*{Acknowledgement}
We thank Jin Liu for his insights and technical assistance, and the strong support and efforts of all creators, managers, and annotators. The work is supported by New Generation Artificial Intelligence-National Science and Technology Major Project (2025ZD0123505), National Natural Science Foundation of China (Grant Nos. 62576338, 62550062, 62425606, 32341009, 62506362), and Beijing Natural Science Foundation (Grant Nos. L252145, L257008).

\section*{Impact Statement}
This research utilizes personal video blog content, which inherently involves sensitive data including individuals' likenesses, personal environments, activities, and associated metadata. We acknowledge the potential privacy risks, such as unauthorized identity exposure or inference of personal habits, especially if such data were to be mishandled or released without safeguards. To mitigate these risks and ensure ethical compliance, our data collection and usage strictly adhere to the following protocols:

\begin{itemize}
    \item \textbf{Data Compliance and Ethical Adherence:} All personal vlog content incorporated into our dataset is collected from public platform under permission of the original creators, in accordance with relevant copyright and license. 

    \item \textbf{Data Anonymization:} During preprocessing, we implement measures to anonymize sensitive elements where feasible, focusing the model’s evaluation on editing structure and narrative quality rather than personal identity.

    \item \textbf{Controlled Usage Scope:} The dataset is used solely for training and benchmarking vlog reward models in this research. It should not be publicly distributed in its raw form to prevent potential misuse.
\end{itemize}

\textbf{Research Implications.} Our contributions support the research community by providing a standardized evaluation criteria and a benchmark for vlog editing assessment. For creators, VlogReward offers an assistive tool to generate detailed, interpretable feedback on editing drafts, potentially lowering barriers to high-quality production and enhancing creative workflows. 

However, automated evaluation systems may inadvertently promote stylistic homogenization or embed biases present in the training data. To mitigate this, we advocate for VlogReward's use as a supportive tool within a human-in-the-loop creative process, not as an autonomous arbiter of quality. The multi-dimensional, feedback-rich output is designed to augment, not replace human judgment. Future work must focus on expanding dataset diversity across cultures and genres, implementing debiasing techniques, and ensuring transparency in model limitations.

\textbf{JSON Editing Plans vs Rendered Vlogs.} We initially compared the 2 approaches in detail, and chose JSON for two primary reasons:
\begin{itemize}
    \item \textbf{Efficiency in Production Scenarios:} In real-world applications like "one-click" vlog creation in editing software, the internal automated evaluation algorithm must quickly select the best among multiple AI-generated plans. Waiting to render each plan into a final video would cause unacceptable delays. Since all real-world editing actions can be represented as structured data, using JSON would be more immediate and efficient.

    \item \textbf{Capturing Fine-Grained Editing Details:} Current multimodal models struggle to accurately perceive dynamic effects in edited vlogs due to their discrete frame-sampling nature. In contrast, a JSON file explicitly documents every editing operation, allowing the model to understand all actions more reliably.
\end{itemize}

\nocite{langley00}

\bibliography{example_paper}
\bibliographystyle{icml2026}

\newpage
\appendix
\onecolumn
\section{Limitations}
\paragraph{Scope of Vlog Types and Generalization of Editing Attributes.} Our work primarily focuses on montage-style vlogs, a dominant but specific genre. The evaluation framework, dataset, and model may not fully capture the unique narrative structures and editing conventions of other popular formats, such as sit-down monologues, detailed tutorials, or highly stylized cinematic pieces. Future work should expand the taxonomy and benchmark to encompass a broader spectrum of vlog genres. Furthermore, the editable attributes in our editing plan representation are standardized to core structural elements (e.g., cut timestamps, shot order, subtitles) to accommodate MLLM processing. This design intentionally excludes fine-grained stylistic controls (e.g., specific transition effects, color grading, complex overlay animations, music) that professional editors frequently manipulate. Extending the model to understand and evaluate these richer attributes is crucial for real-world applications.

\vspace{-6pt}
\paragraph{Inherent Subjectivity and the ``Ground Truth'' Challenge.} Evaluating creative work is fundamentally subjective. Our six-dimensional taxonomy, though developed with experts, cannot fully codify human artistic judgment. The ``ground truth'' scores in VRMBench, while carefully curated, represent a consensus rather than an absolute standard. The model’s performance is therefore measured against this specific consensus, which may not align with all individual viewers' or creators' perceptions. The task itself has no single absolutely correct answer.
\vspace{-6pt}

\paragraph{Visual Processing Constraints.} To ensure training efficiency, our model processes videos at a low frame rate (1 fps) and a limited resolution. This coarse visual representation may cause the model to miss subtle cinematic details—such as precise motion, fleeting expressions, or nuanced lighting, that are critical for a professional evaluation of cinematography and pacing. This represents a trade-off between computational feasibility and granular visual understanding. Nevertheless, all our experiments were conducted under identical parameter constraints, thereby ensuring fairness in the evaluation and validating the effectiveness of our proposed inter-group comparison reward.

\section{More Detailed Implementation Settings}
\label{appendix implementation}
In this section, we provide more details about our implementations. 

For GRPO training, we optimize the current policy model $\pi_\theta$ by maximizing the following objective
\begin{equation}
J(\theta) 
= \mathbb{E}_{q, \{o_i\}_{i=1}^G \sim \pi_{\theta_{\mathrm{old}}}(O \mid q)} 
\Biggl[
  \frac{1}{G} \sum_{i=1}^G
  \min\!\Bigl(
    \frac{\pi_{\theta}(o_i \mid q)}{\pi_{\theta_{\mathrm{old}}}(o_i \mid q)}\,A_i,\, 
    \mathrm{clip}\!\Bigl(
      \frac{\pi_{\theta}(o_i \mid q)}{\pi_{\theta_{\mathrm{old}}}(o_i \mid q)},
      1-\varepsilon,\,
      1+\varepsilon
    \Bigr)
    A_i
  \Bigr)  
  -\,\beta\,D_{\mathrm{KL}}\bigl(\pi_{\theta}\,\big\|\,\pi_{\mathrm{ref}}\bigr)
\Biggr],
\end{equation}
where $\varepsilon$ and $\beta$ are the clipping hyper-parameter and coefficient controlling the Kullback–Leibler (KL)
penalty \citep{schulman2017proximal}, respectively. $A_i = \frac{\reward_i - \operatorname{mean}(\{\reward_1, \reward_2, \dots, \reward_G\})}{\operatorname{std}(\{\reward_1, \reward_2, \dots, \reward_G\})}$ is the computed relative advantage using the intra-group rewards $\{\reward_1,\reward_2,\cdots,\reward_G\}$ and $D_{KL}(\pi_\theta \,\|\, \pi_{\mathrm{ref}})\!
=\! \frac{\pi_{\mathrm{ref}}(o_i \mid q)}{\pi_\theta(o_i \mid q)}
\!-\! \log\!\Bigl(\frac{\pi_{\mathrm{ref}}(o_i \mid q)}{\pi_\theta(o_i \mid q)}\Bigr)
\!-\! 1$ is the KL divergence.

For RRM-MSE, we replace the LLM head with a six‑dimensional linear output head, whereas RRM‑BT requires only a single‑dimensional linear head. For a fair comparison, we train RRM‑MSE on VRM‑Dataset‑SFT and VRM‑Dataset‑OB, and use the rounded score value for test. For RRM‑BT, we construct training pairs from each group in VRM‑Dataset‑RL, yielding six pairwise comparisons per raw‑footage group. We further explore different learning rates for both RRM‑MSE and RRM‑BT, whose results are summarized in Table \ref{tab differnet lr}, respectively. We report the relatively best achieved performance in Table \ref{tab1}. The detailed configuration settings of SFT, GRPO, RRM-MSE and RRM-BT are summarized in Table \ref{tab paras}.

\begin{table}
    \centering
    \caption{Performances of RRM-MSE and RRM-BT using different learing rates.}
    \begin{tabular}{l|c|ccc|ccc|ccc}
    \toprule
    \multirow{2}{*}{\textbf{Models}} &\multirow{2}{*}{\textbf{Learning Rate}} & \multicolumn{3}{c|}{\textbf{Score Accuracy}} & \multicolumn{3}{c|}{\textbf{Comparison Accuracy}} & \multicolumn{3}{c}{\textbf{Best-of-N Accuracy}} \\
        \cmidrule{3-11}
        & & SB & OB & AVG & SB & OB & AVG & SB & OB & AVG \\
        \midrule
     \multirow{7}{*}{RRM-MSE}  &1e-4  &\textbf{25.4} 	&23.3 	&24.4 	&35.5 	&26.5 	&31.0 	&25.8 	&24.6 	&25.2   \\
       &1e-3  &24.8 	&25.9 	&25.4 	&38.1 	&32.2 	&35.1 	&31.1 	&28.9 	&30.0\\ 
       &1e-2  &23.6 	&25.4 	&24.5 	&42.3 	&45.5 	&43.9 	&\textbf{31.6} 	&41.0 	&\textbf{36.3} \\
       &1e-1  &23.4 	&\textbf{29.8} 	&\textbf{26.6} 	&43.0 	&44.8 	&43.9 	&28.0 	&37.5 	&32.8 \\
       &2e-1  &18.5 	&18.1 	&18.3 	&41.9 	&46.3 	&44.1 	&24.9 	&38.2 	&31.6 \\
       &5e-1  &4.5 	&2.3 	&3.4 	&44.6 	&\textbf{55.8} 	&\textbf{50.2 }	&25.2 	&\textbf{45.2} 	&35.2 \\
       &1    &3.4 	    &3.4 	&3.4 	&\textbf{47.5} 	&39.8 	&43.7 	&27.9 	&23.2 	&25.6 \\

        \midrule
      \multirow{7}{*}{RRM-BT}  &1e-4 &- &-&-  &44.2 	&70.1 	&57.1 	&24.5 	&49.4 	&37.0 \\
               &1e-3  &- &-&-  &53.0 	&79.8 	&66.4 	&29.0 	&63.8 	&46.4 \\ 
               &1e-2  &- &-&-  &56.7 	&82.6 	&69.6 	&35.0 	&\textbf{72.4 }	&\textbf{53.7} \\ 
               &1e-1  &- &-&-  &58.4 	&\textbf{83.1} 	&\textbf{70.8} 	&36.5 	&69.5 	&53.0 \\ 
               &2e-1  &- &-&-  &\textbf{58.7 }	&81.8 	&70.2 	&35.7 	&70.0 	&52.9 \\ 
               &5e-1  &- &-&-  &56.3 	&78.9 	&67.7 	&\textbf{40.3} 	&61.5 	&50.9 \\ 
               &1     &- &-&-  &58.6 	&82.8 	&70.7 	&33.0 	&68.1 	&50.6 \\ 
    \bottomrule
    \end{tabular}
    
    \label{tab differnet lr}
    
\end{table}

\begin{table}
\centering
\caption{Detailed configuration settings of SFT, GRPO, RRM-MSE and RRM-BT.}
\begin{tabular}{l | c c c c}
\toprule
\textbf{Configuration}       & \textbf{SFT} & \textbf{GRPO} & \textbf{RRM-MSE} & \textbf{RRM-BT} \\
\midrule
freeze\_visual\_encoder      & \multicolumn{4}{c}{True}         \\
tune\_merger  & \multicolumn{4}{c}{False} \\
freeze\_llm   &False  &False &True &True                  \\
learning\_rate               & 2e-6             & 1e-6             & 1e-2             & 1e-1             \\
kl\_loss\_coef ($\beta$)     & -                & 1e-2             & -                & -             \\
rollout\_number & - &8 &- &- \\
rollout\_temperature &- & 1 &- &- \\
inference\_temperature &0.01 &0.01 &-&- \\
inference\_top\_p &0.001 &0.001 & - &- \\
optimizer                    &\multicolumn{4}{c}{AdamW}                  \\
AdamW\_betas                 & \multicolumn{4}{c}{ (0.9, 0.999)}             \\
weight\_decay                & 0              & 1e-2             & 0              & 0                \\
warmup\_ratio                & 0.03              & 0                & 0.05            & 0.05                \\
lr\_scheduler                &\multicolumn{4}{c}{cosine}         \\
group\_size                  &-      & 8                & -    & -                \\
batch\_size                  & 128              & 64               & 128              & 128               \\
number\_of\_epochs           & 2                & 1                & 2                & 2                     \\
max\_model\_length        & 16384             & 16384             & 6144            & 6144            \\
max\_total\_pixels           & \multicolumn{4}{c}{64$\times$28$\times$28}\\
fps                 &\multicolumn{4}{c}{1}         \\ 
min\_frames          &\multicolumn{4}{c}{16}         \\      
max\_frames          &\multicolumn{4}{c}{64}         \\
sample\_num &100k &40k &120k &60k \\ 
\midrule
\multirow{2}{*}{Dataset}  &\multirow{2}{*}{VRM-Dataset-SFT}  &\multirow{2}{*}{VRM-Dataset-RL}  &VRM-Dataset-SFT &\multirow{2}{*}{VRM-Dataset-RL} \\
& & & VRM-Dataset-OB \\

\bottomrule
\end{tabular}
\label{tab paras}
\end{table}

\section{Experiments Results Using Thinking Mode}
\label{appendix thinking mode}
We don't use R1-like output mode with \texttt{<think>...</think>} and \texttt{<answer>...</answer>}, as we find it does not improve the performance, yet significantly increasing the overhead. Similar to R1-series work, we set the system prompt as follows when testing the performance in this thinking mode.

\texttt{A conversation between User and Assistant. The user asks a question, and the assistant solves it. The assistant first thinks about the reasoning process in the mind and then provides the user with the answer. The reasoning process and answer are enclosed within <think> </think> and <answer> </answer> tags, respectively, i.e., <think> reasoning process here </think><answer> answer here </answer>}.

Experiment results are reported in Table \ref{tab:thinking mode}. We can see that GRM-GRPO and VlogReward using thinking mode are comparable to or even worse in overall performance than those without it. This suggests that the thinking mode is not always suitable for all tasks. It may be because that vlog evaluation primarily relies on immediate perceptual judgment rather than extended logical reasoning, making the verbose thinking process potentially redundant or distracting.

\begin{table}
  \centering
  \vspace{10pt}
  \caption{Accuracy performances of GRM-GRPO and VlogReward with and without employing the thinking mode.}
  \label{tab:thinking mode}
  \begin{tabular}{l|l|ccc|ccc|ccc}
    \toprule
    \multirow{2}{*}{\textbf{Models}}     &\multirow{2}{*}{\textbf{Mode } }     & \multicolumn{3}{c|}{\textbf{Score Accuracy}} & \multicolumn{3}{c|}{\textbf{Comparison Accuracy}} & \multicolumn{3}{c}{\textbf{Best-of-N Accuracy}} \\
    \cmidrule{3-11}
              &            & SB   & OB    & AVG   & SB    & OB    & AVG   & SB    & OB    & AVG   \\
    \midrule
    GRM-GRPO   & no thinking & \textbf{44.0}   & \textbf{56.5}  & \textbf{50.3}  & 54.8  & \textbf{84.0}    & \textbf{69.4}  & \textbf{43.7}  & \textbf{86.2}  & \textbf{65.0 }   \\
  
    GRM-GRPO   & thinking   & 43.8 & 55.7  & 49.7  & \textbf{56.9}  & 81.8  & \textbf{69.4}  & 43.0  & 82.9  & \textbf{63.0}  \\
    \midrule
    VlogReward & no thinking & \textbf{45.9} & 56.7  & \textbf{51.3}  & 60.4  & 86.6  & 73.5  & \textbf{48.3}  & 86.8  & \textbf{67.6 } \\
    
    VlogReward & thinking   & 44.4 & \textbf{57.2}  & 50.8  & \textbf{61.6}  & \textbf{88.2}  & \textbf{74.9}  & 43.7  & \textbf{88.1}  & 65.9  \\
    \bottomrule
  \end{tabular}
\end{table}

\section{Data Examples}

\begin{figure}[!h]
    \centering
    \includegraphics[width=0.9\linewidth]{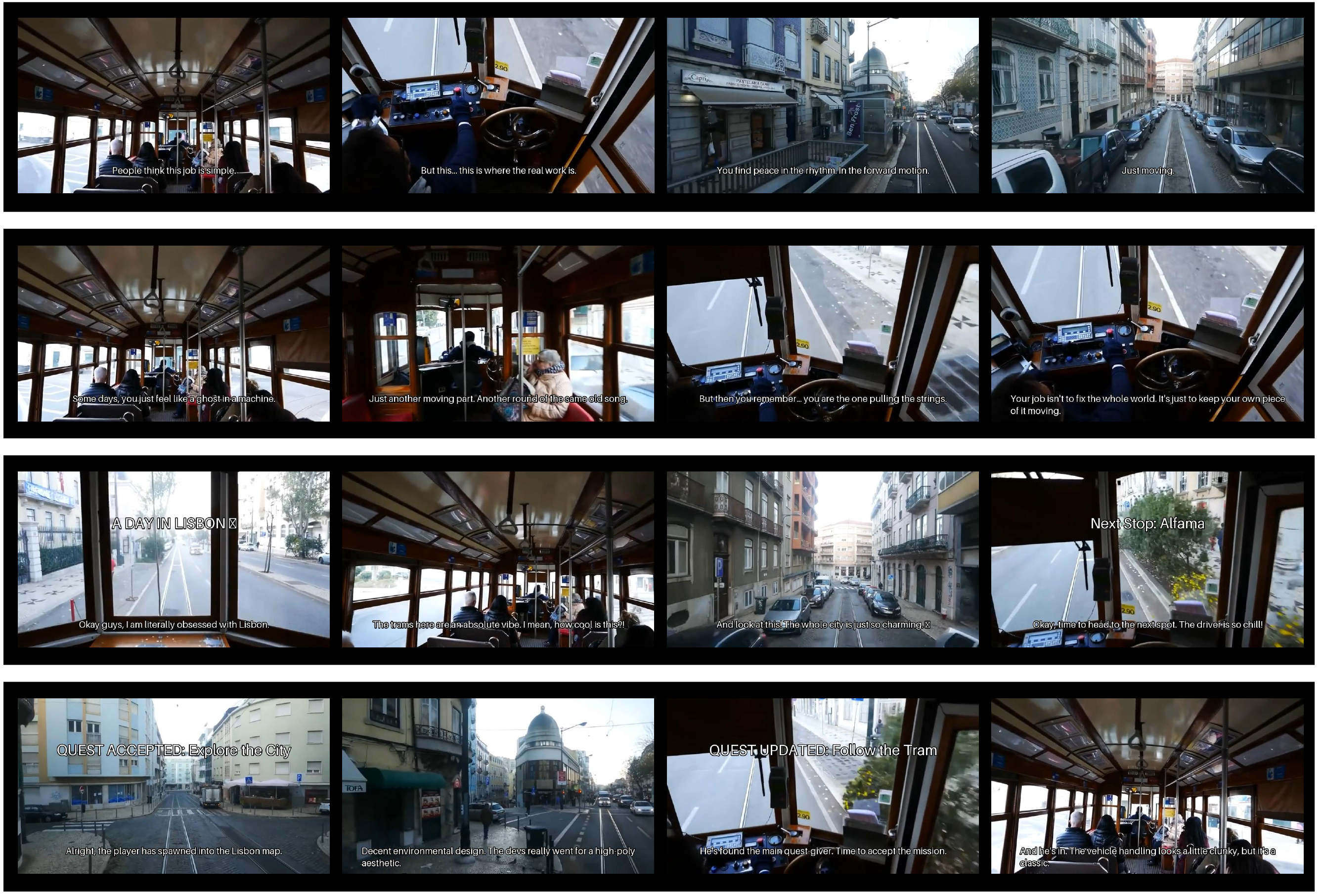}
    \caption{Examples of edited vlogs in the subjective dataset. Here we show some segments of the final vlogs edited with corresponding editing plans for the convenience of review. The vlog script topics are various such as reflections on job and life, urban travel, and game-style narratives.}
    \label{fig:ob data example}
\end{figure}

\begin{figure}[!h]
    \centering
    \includegraphics[width=0.9\linewidth]{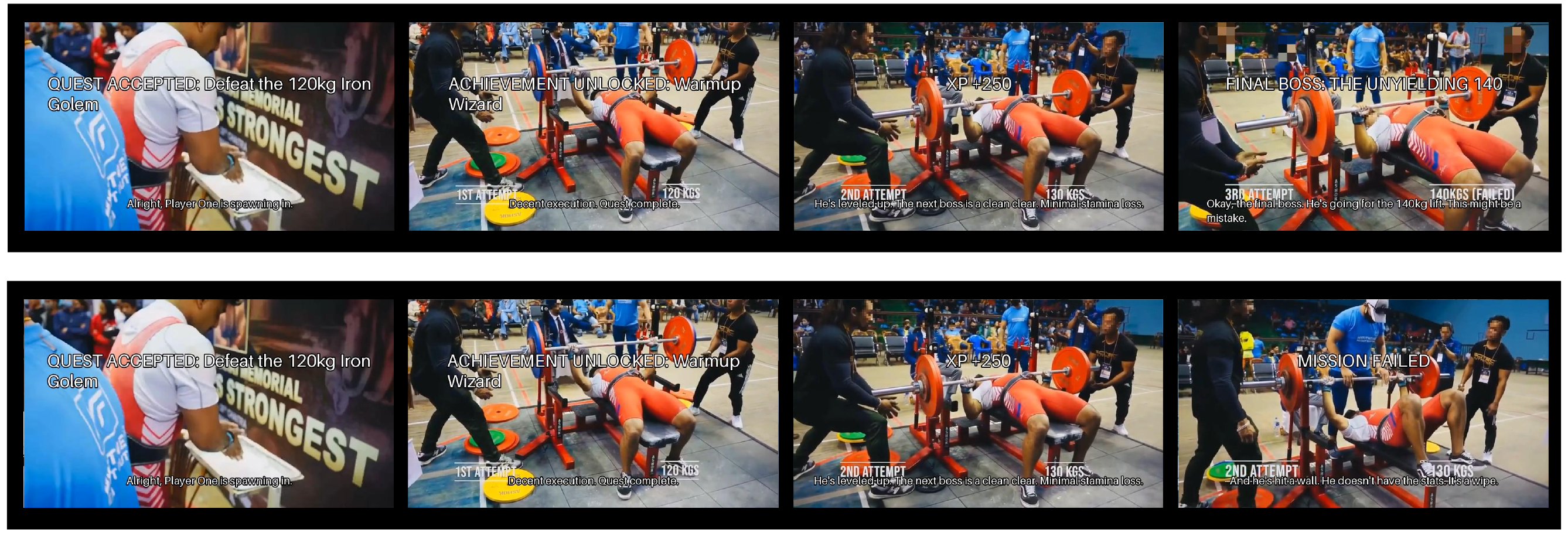}
    \caption{Examples of edited vlogs in the objective dataset. The vlog at the bottom is obtained by tampering with the timestamps of the vlog above, introducing temporal and descriptive errors.
    }
    \label{fig:ob data example}
\end{figure}

\newpage
\section{Evaluation Criteria Details}
\label{appendix criteria}
Under the guidance of professional vlog creators and product managers, we have established a comprehensive taxonomy of six dimensions to evaluate vlog editing plans. Each dimension is assessed on a 1-5 integer scale: 5 for outstanding, 4 for excellent, 3 for acceptable, 2 for poor, and 1 for failure. The specific evaluation emphasis for the six dimensions are summarized as follows. 

\begin{itemize}
    \item \textbf{Creativity \& Originality:} Evaluates narrative inventiveness, unique perspectives, and the presence of generic clichés.

    \item \textbf{Script \& Visual Plan Consistency:} Measures how well the selected visual clips match the specific title, subtitles and voiceover.

    \item \textbf{Concept Design:} Assesses the appropriateness of script elements, theme logic, and the cohesive portrayal of characters and settings.

    \item \textbf{Cinematography \& Clip Selection:} Evaluates the aesthetic quality of shot selection and framing, as well as the precision in picking the most expressive segments from raw assets.

    \item  \textbf{Narration \& Voice-Over:} Focuses on the quality of the storyline, scriptwriting depth, and the emotional delivery of the narration.

    \item  \textbf{Pacing \& Rhythm:} Evaluates the temporal rhythm, sequence logic, and the overall chronological flow of the edit.
\end{itemize}

When training reward models, the input prompt does not detail the scoring criteria for each dimension. In contrast, when evaluating other baselines that are not trained on VRMDataset, the input prompt explicitly defines the following scoring rubrics in detail.

\clearpage
\begin{tcolorbox}[
    colback=gray!5, 
    colframe=blue!80!black,
    title=\textbf{Creativity \& Originality}, 
    fontupper=\ttfamily,
    boxrule=0.5mm,
    arc=2mm, 
    bottom=2mm, top=2mm,
    fonttitle=\fontsize{12}{16}\selectfont\bfseries,
]
\label{xxx}
\begin{scorebox}{Score 5 (Outstanding)}
\textbf{Visionary and masterfully structured.} The narrative is deeply engaging, exceptionally creative, and feels fresh and memorable. It represents a perfect fusion of idea and execution, offering unique insights.
\end{scorebox}

\begin{scorebox}{Score 4 (Excellent)}
\textbf{Well-crafted and compelling.} The story has a clear, logical structure and a coherent, engaging plot. The language is precise and effectively serves the narrative, even if it doesn't break new creative ground.
\end{scorebox}

\begin{scorebox}{Score 3 (Acceptable)}
\textbf{Functional but lacks impact.} The narrative feels predictable or unengaging, often due to a \emph{flat plot} or \emph{loose structure}. The story is understandable but fails to leave a lasting impression.
\end{scorebox}

\begin{scorebox}{Score 2 (Poor)}
\textbf{Illogical and confusing.} Key narrative elements are missing or poorly explained, leaving the audience with more questions than answers. The primary issue is a \emph{lack of clear logic}.
\end{scorebox}

\begin{scorebox}{Score 1 (Failure)}
\textbf{Chaotic and incoherent.} The story completely lacks a discernible plot or structure. The content is irrelevant to the stated theme. The primary issue is a \emph{chaotic or non-existent plot}.
\end{scorebox}

\end{tcolorbox}

\begin{tcolorbox}[
    colback=gray!5, 
    colframe=blue!80!black,
    title=\textbf{Script \& Visual Plan Consistency}, 
    fontupper=\ttfamily,
    boxrule=0.5mm,
    arc=2mm, 
    bottom=2mm, top=2mm,
    fonttitle=\fontsize{12}{16}\selectfont\bfseries
]
\begin{scorebox}{Score 5 (Outstanding)}
\textbf{Perfect synergy.} The visuals not only match but elevate the script's message and emotional intent. Every frame feels intentional and deeply connected to the narrative, creating a synergistic masterpiece. The plan is \emph{perfectly consistent}.
\end{scorebox}

\begin{scorebox}{Score 4 (Excellent)}
\textbf{Consistently effective.} The visual plan strongly supports the script. Nearly all visual choices are well-justified and enhance the narrative with no significant contradictions.
\end{scorebox}

\begin{scorebox}{Score 3 (Acceptable)}
\textbf{Partially disconnected.} A noticeable gap exists where the visuals don't fully align with the script's core message or tone. For example, the script describes excitement, but the shots are static and low-energy. The main storyline, however, remains comprehensible.
\end{scorebox}

\begin{scorebox}{Score 2 (Poor)}
\textbf{Frequently contradictory.} The visual plan often undermines or fails to support the script's intent, creating confusion and weakening the story's impact.
\end{scorebox}

\begin{scorebox}{Score 1 (Failure)}
\textbf{Severe conflict.} The visuals are so disconnected from the script that they fundamentally sabotage the intended message. The primary issue is a \emph{conflicting message}.
\end{scorebox}
\end{tcolorbox}

\begin{tcolorbox}[
    colback=gray!5, 
    colframe=blue!80!black,
    title=\textbf{Concept \& Asset Design}, 
    fontupper=\ttfamily,
    boxrule=0.5mm,
    arc=2mm, 
    bottom=2mm, top=2mm,
    fonttitle=\fontsize{12}{16}\selectfont\bfseries
]
\begin{scorebox}{Score 5 (Outstanding)}
\textbf{Impeccably immersive world-building.} The set design, props, and character roles are meticulously planned and seamlessly integrated to create a believable and authentic environment that enhances the story. Strengths: \emph{flawless set design}, \emph{purposeful props}, \emph{authentic character roles}.
\end{scorebox}

\begin{scorebox}{Score 4 (Excellent)}
\textbf{Cohesive and well-chosen.} The set design, props, and other assets create a consistent and believable world that supports the narrative. All key assets are present and used logically.
\end{scorebox}

\begin{scorebox}{Score 3 (Acceptable)}
\textbf{Inconsistent.} Some conceptual elements, character roles, or props feel out of place or conflict with the storyline. For example, a character meant to be a medieval knight is shown using a smartphone. Key issues: \emph{illogical role settings} or \emph{inconsistent scene/prop usage}.
\end{scorebox}

\begin{scorebox}{Score 2 (Poor)}
\textbf{Confusing and poorly chosen.} The conceptual assets are mismatched, creating a nonsensical visual experience that actively detracts from the story's credibility.
\end{scorebox}

\begin{scorebox}{Score 1 (Failure)}
\textbf{Critical assets are missing or illogical.} Key character roles or props are either absent entirely or so inappropriate that they cause a total breakdown in narrative logic. Key issues: \emph{missing key roles/props}.
\end{scorebox}
\end{tcolorbox}

\begin{tcolorbox}[
    colback=gray!5, 
    colframe=blue!80!black,
    title=\textbf{Cinematography \& Clip Selection}, 
    fontupper=\ttfamily,
    boxrule=0.5mm,
    arc=2mm, 
    bottom=2mm, top=2mm,
    fonttitle=\fontsize{12}{16}\selectfont\bfseries
]
\begin{scorebox}{Score 5 (Outstanding)}
\textbf{Visionary and precise.} The cinematography is visually stunning (composition, movement, lens choice). The clip selection is \emph{perfectly timed} to capture the absolute peak of action, emotion, or narrative importance. Every selected segment is the most impactful choice from the raw footage. Strengths: \emph{masterful cinematography}, \emph{precise and impactful clip selection}.
\end{scorebox}

\begin{scorebox}{Score 4 (Excellent)}
\textbf{Professional and effective.} The cinematography is well-executed, and the clip selection is logical and accurate. The chosen segments effectively convey the intended action and narrative without significant omissions.
\end{scorebox}

\begin{scorebox}{Score 3 (Acceptable)}
\textbf{Functional but suboptimal.} The cinematography is uninspired (e.g., \emph{simple shot composition}, \emph{lack of dynamism}). The clip selection is relevant but often \emph{misses the key moments}, trimming a shot too early or letting it run too long, weakening its impact.
\end{scorebox}

\begin{scorebox}{Score 2 (Poor)}
\textbf{Technically weak selection. }Shots are frequently out of focus or poorly framed. The chosen clips often obscure essential information or focus on irrelevant action, failing to capture the core of the moment.
\end{scorebox}

\begin{scorebox}{Score 1 (Failure)}
\textbf{Chaotic and illogical.} The camera work is incoherent, and the chosen clips are either completely irrelevant to the script or trimmed so poorly that they make no sense, resulting in a baffling visual experience.
\end{scorebox}
\end{tcolorbox}

\begin{tcolorbox}[
    colback=gray!5, 
    colframe=blue!80!black,
    title=\textbf{Narration \& Voice-Over}, 
    fontupper=\ttfamily,
    boxrule=0.5mm,
    arc=2mm, 
    bottom=2mm, top=2mm,
    fonttitle=\fontsize{12}{16}\selectfont\bfseries
]
\begin{scorebox}{Score 5 (Outstanding)}
\textbf{Captivating and flawless.} The writing is exceptional, and the delivery is emotionally resonant with a distinctive style. The narration adds profound thematic depth and elevates the entire production. Strengths: \emph{brilliant writing}, \emph{rich emotional delivery}, \emph{deepened thematic meaning}.
\end{scorebox}

\begin{scorebox}{Score 4 (Excellent)}
\textbf{Clear and professional.} The narration effectively tells the story with fluent language and a consistent tone that perfectly matches the content. The audio is clean and well-mixed.
\end{scorebox}

\begin{scorebox}{Score 3 (Acceptable)}
\textbf{Basic but functional.} The content may be superficial, the language bland, or the delivery flat and unemotional. Key issues: \emph{empty content}, \emph{clumsy wording}, or \emph{narration that doesn't fit the mood}.
\end{scorebox}

\begin{scorebox}{Score 2 (Poor)}
\textbf{Detracts from the experience.} The narration is filled with clichés, awkward phrasing, or is poorly matched with the visuals, actively weakening the final product.
\end{scorebox}

\begin{scorebox}{Score 1 (Failure)}
\textbf{Irrelevant or nonsensical.} The narration is completely disconnected from the story, contains significant factual errors, or is otherwise incomprehensible.
\end{scorebox}
\end{tcolorbox}

\begin{tcolorbox}[
    colback=gray!5, 
    colframe=blue!80!black,
    title=\textbf{Pacing \& Rhythm}, 
    fontupper=\ttfamily,
    boxrule=0.5mm,
    arc=2mm, 
    bottom=2mm, top=2mm,
    fonttitle=\fontsize{12}{16}\selectfont\bfseries
]
\begin{scorebox}{Score 5 (Outstanding)}
\textbf{Masterful and dynamic.} The editing has a flawless, instinctive rhythm. The pacing seamlessly blends shots, sound, and story beats to create a compelling and captivating viewing experience. The use of all assets is \emph{perfectly integrated}.
\end{scorebox}

\begin{scorebox}{Score 4 (Excellent)}
\textbf{Smooth and effective.} The editing is well-paced, and the flow logically guides the viewer's attention and supports the narrative's emotional arc. Transitions are purposeful and support the story.
\end{scorebox}

\begin{scorebox}{Score 3 (Acceptable)}
\textbf{Inconsistent.} The editing pace feels uneven, with some sections dragging or feeling rushed. This occasionally disrupts the narrative flow, but the overall story remains understandable. Transitions might be abrupt or cliché.
\end{scorebox}

\begin{scorebox}{Score 2 (Poor)}
\textbf{Jarring and amateurish.} Cuts are poorly timed, transitions are awkward, and the overall rhythm feels chaotic, frequently pulling the viewer out of the experience.
\end{scorebox}

\begin{scorebox}{Score 1 (Failure)}
\textbf{Critically flawed. }The editing has a \emph{chaotic flow} that completely disrupts the story's logic. Illogical cuts or asset usage make the narrative incomprehensible.
\end{scorebox}
\end{tcolorbox}

\newpage
\begin{tcolorbox}[
    colback=gray!5, 
    colframe=blue!70!black,
    title=\textbf{Vlog Rewarding Prompt}, 
    fontupper={\fontsize{8.8}{8.5}\selectfont\ttfamily},
    boxrule=0.5mm,
    arc=2mm, 
    bottom=2mm, top=2mm,
    fonttitle=\fontsize{12}{16}\selectfont\bfseries
]

\section*{Your Role}

You are an expert video editor and creative director. Your task is to provide a rigorous and constructive evaluation of a vlog editing plan. You will be given a video file containing all raw clips and a detailed editing script. You must score the plan from 1 to 5 across six core dimensions and provide specific, actionable feedback for each score. Your feedback must clearly explain what works and what does not, referencing specific timestamps from the script or raw video file.

\section*{Input}

\begin{enumerate}[label=\textbf{\arabic*.}]
    \item \textbf{A complete video file:} A single video file containing a sequence of all raw clips, stitched together with a 3-second black screen separating each clip.
    
    \item \textbf{A detailed vlog editing plan:} A list containing the selected segments, detailing the cut timestamps, script/narration, shot order, visual effects, and other key editing decisions. If the duration of a segment's \texttt{shot\_timestamp} doesn't match its assigned \texttt{final\_timestamp}, this means adjusting its playback speed (speed up or slow down) so its length can match the required \texttt{final\_timestamp} in the edit.
\end{enumerate}

\section*{Scoring Dimensions and Rubric}

\subsection*{1. Creativity \& Originality}

\subsection*{2. Script \& Visual Plan Consistency}

\subsection*{3. Concept \& Asset Design}

\subsection*{4. Cinematography \& Clip Selection}

\subsection*{5. Narration \& Voice-Over}

\subsection*{6. Pacing \& Rhythm}

\section*{Your Final Task}
For each of the 6 dimensions, provide a score from 1 to 5 and deliver specific, constructive feedback. Justify every score with clear examples. For instance, instead of saying, ``The pacing is off,'' say, ``In the sequence from 0:45 to 1:10, the editing lingers on wide shots for too long, causing the energy to drop. The script implies rising tension, which could be better achieved with quicker cuts between the close-ups in the raw footage.''

\section*{Output Format}

After reasoning, provide a \emph{plain text} evaluation with scores and feedback for each aspect.

\subsection*{Output Example}

\begin{verbatim}
1. Creativity & Originality
Score: 2
Feedback: ...

2. Script & Visual Plan Consistency
Score: 1
Feedback: ...

3. Concept & Asset Design
Score: 4
Feedback: ...

4. Cinematography & Clip Selection
Score: 3
Feedback: ...

5. Narration & Voice-Over
Score: 3
Feedback: ...

6. Pacing & Rhythm
Score: 1
Feedback: ...
\end{verbatim}

\section*{Vlog Editing Plan}

\{vlog\_editing\_plan\}
\end{tcolorbox}



\end{document}